\documentclass[11pt]{article}

\usepackage[preprint]{acl}

\usepackage{times}
\usepackage{latexsym}

\usepackage[T1]{fontenc}
\usepackage[utf8]{inputenc}

\usepackage{microtype}

\usepackage{inconsolata}

\usepackage{graphicx}

\title{BEFT: Bias-Efficient Fine-Tuning of Language Models\\ \aclbaichuan{in Low-Data Regimes}}

\author{
 \textbf{Baichuan Huang\textsuperscript{1}},
 \textbf{Ananth Balashankar\textsuperscript{2}\thanks{contributed in an advisory role.}},
 \textbf{Amir Aminifar\textsuperscript{1}},
\\
 \textsuperscript{1}Lund University, Sweden,
 \textsuperscript{2}Google DeepMind, USA,
\\
\{baichuan.huang, amir.aminifar\}@eit.lth.se, ananthbshankar@google.com
}

\usepackage[acronym]{glossaries}
\usepackage{pifont}
\usepackage{xspace}
\usepackage{subcaption}
\usepackage[most]{tcolorbox}
\usepackage{booktabs}
\usepackage{subcaption}  
\usepackage{multirow}
\usepackage{multicol}
\usepackage{tikz}
\usepackage{textcomp}

\newacronym{FO}{FO}{first-order}
\newacronym{ZO}{ZO}{zeroth-order}
\newacronym{SGD}{SGD}{stochastic gradient descent}
\newacronym{LLM}{LLM}{large language model}
\newacronym{LoRA}{LoRA}{low-rank adaptation}
\newacronym{BP}{BP}{Backpropagation}
\newacronym{IP}{IP}{in-place}
\newacronym{PEFT}{PEFT}{parameter-efficient fine-tuning}
\newacronym{FT}{FT}{fine-tuning}
\newacronym{MeZO}{MeZO}{memory-efficient zeroth-order optimizer}
\newacronym{CPU}{CPU}{central processing unit}
\newacronym{GPU}{GPU}{graphics processing unit}
\newacronym{RAM}{RAM}{random-access memory}
\newacronym{SPSA}{SPSA}{Simultaneous Perturbation Stochastic Approximation}
\newacronym{BEFT}{BEFT}{bias-efficient fine-tuning}
\newacronym{ICL}{ICL}{in-context learning}
\newacronym{seconds}{s}{seconds}
\newacronym{VeRA}{VeRA}{vector-based random matrix adaptation}
\newacronym{DoRA}{DoRA}{weight-decomposed low-rank adaptation}
\newacronym{SDPA}{SDPA}{scaled dot-product attention}

\newcommand{\boldeq}{\ensuremath{\boldsymbol}} %for math
\usepackage{booktabs} 
\usepackage{amssymb}
\usepackage{amsmath}

\newcommand{\baichuanchange}{\textcolor{black}}
\newcommand{\aclbaichuan}{\textcolor{black}}
\newcommand{\camera}{\textcolor{black}}
\glsdisablehyper

\usepackage{listings}
\usepackage{xcolor}

\usepackage[table]{xcolor}
\definecolor{mygreen}{HTML}{55A868}
\definecolor{myred}{HTML}{C44F51}
\definecolor{myblue}{HTML}{4C72B0}

\usepackage{float}
\usepackage{caption}

\usepackage{fancyhdr}
\begin{document}
\pagestyle{fancy}
\fancyhf{}
\lhead{\footnotesize \textcolor{black}{Published as a main conference paper at ACL 2026}}

\maketitle
\begin{abstract}
Fine-tuning the bias terms of \glspl{LLM} has the potential to achieve unprecedented parameter efficiency while maintaining competitive performance, particularly in low-data regimes. However, the link between fine-tuning different bias terms (i.e., $\boldeq{b}_q$, $\boldeq{b}_k$, and $\boldeq{b}_v$ in the query, key, or value projections) and downstream performance remains largely unclear to date. In this paper, we investigate the link between fine-tuning $\boldeq{b}_q$, $\boldeq{b}_k$, and $\boldeq{b}_v$ with the performance of the downstream task. 
Our key finding is that \textit{directly fine-tuning $\boldeq{b}_v$ generally leads to higher downstream performance in low-data regimes, in comparison to $\boldeq{b}_q$ and $\boldeq{b}_k$.} We extensively evaluate this unique property across a wide range of \glspl{LLM} spanning encoder-only and decoder-only architectures up to 6.7B parameters (including bias-free \glspl{LLM}). Our results provide strong evidence for the effectiveness of directly fine-tuning $\boldeq{b}_v$ across various downstream tasks. The implementation code is available at \url{https://github.com/whubaichuan/BEFT}.
\end{abstract}

\glsresetall

\section{Introduction}

Fine-tuning pre-trained \glspl{LLM} for downstream tasks has gained a lot of attention over the past few years. \Gls{PEFT} methods have been widely studied \cite{ding2023parameter,yang2024harnessing} to reduce the fine-tuning overheads, such as computation cost, \gls{GPU} memory usage, and energy consumption. Among these \gls{PEFT} techniques, bias-only fine-tuning—which involves updating only the bias terms of the \glspl{LLM} \cite{zaken2022bitfit}—provides the potential for unprecedented parameter efficiency.

Bias-only fine-tuning offers out-of-the-box usability, without extra requirements for additional configuration and auxiliary reparameterization. This is in contrast to the majority of the previous work in this domain, e.g., prefix tuning \cite{li2021prefix}, adapter tuning \cite{houlsby2019parameter}, LoRA \cite{hu2022lora}, and GaLore \cite{zhao2024galore}. Moreover, in low-data regimes, fine-tuning all bias terms has been shown to be competitive with full-parameter fine-tuning, despite updating a significantly smaller subset of full parameters \cite{zaken2022bitfit,logan2022cutting}. Despite the advantages of bias-only fine-tuning, the \baichuanchange{relationship} between \baichuanchange{fine-tuning different bias terms} and downstream performance remains largely unclear \cite{ding2023parameter,wengbitfit+}.

\begin{figure}[!t]
    \hspace*{-0.7cm}
    \includegraphics[width=1.1\linewidth]{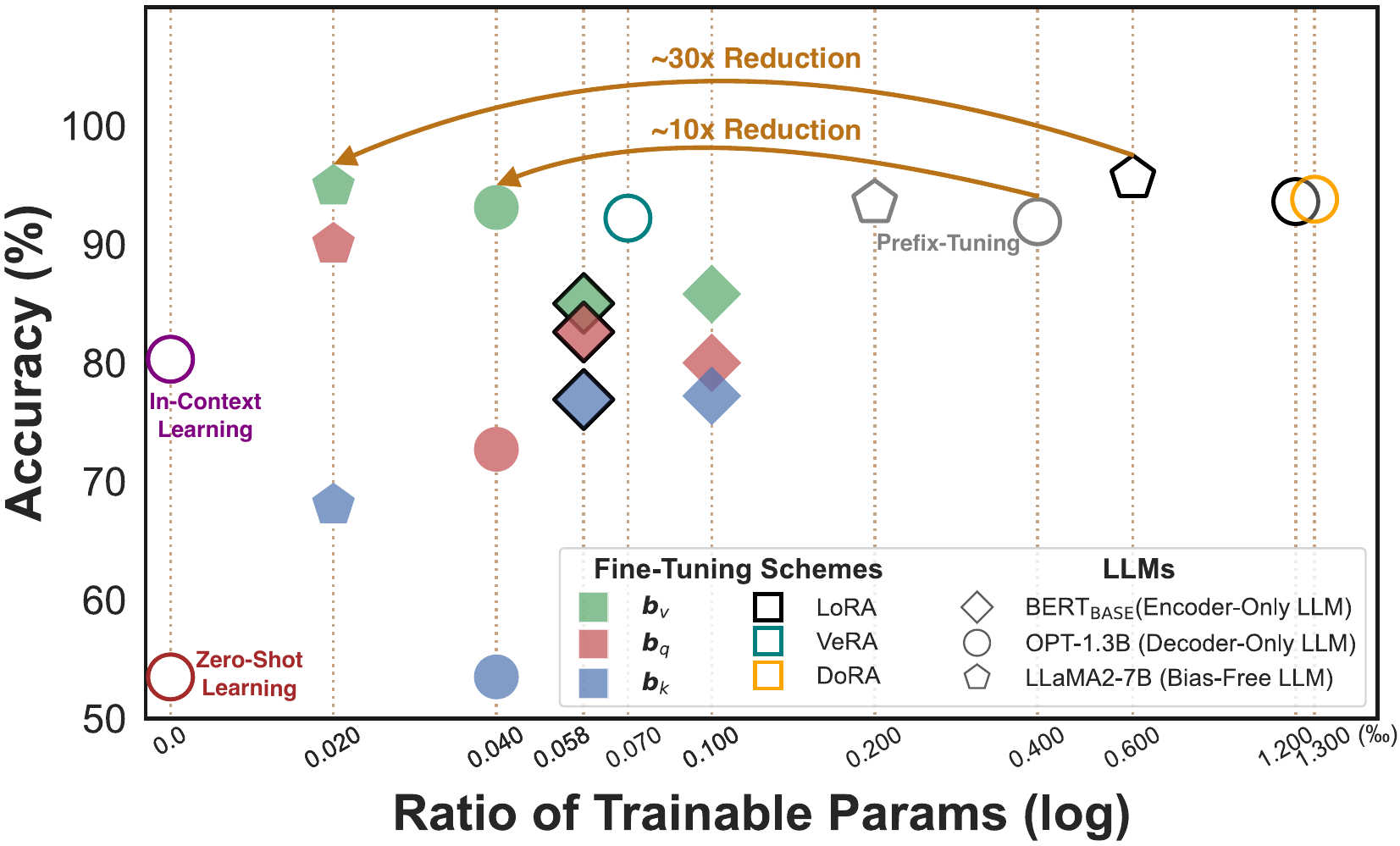}
  \caption{Fine-tuning $\boldeq{b}_v$ (green) leads to higher downstream performance, compared to $\boldeq{b}_q$ (red) and $\boldeq{b}_k$ (blue) (even combined with \camera{\gls{PEFT} such as LoRA}), on the SST-2 dataset with low-data regime ($1000$ training samples).}
  \label{fig:observation_new}
  \vspace{-10pt}
\end{figure}

In this paper, we investigate the link between fine-tuning different bias terms (i.e., $\boldeq{b}_q$, $\boldeq{b}_k$, and $\boldeq{b}_v$ in the query, key, or value projections) and downstream performance. 
We observe that fine-tuning $\boldeq{b}_v$ generally leads to higher downstream performance in low-data regimes, compared to $\boldeq{b}_q$ and $\boldeq{b}_k$. For instance, for BERT$_{\mathrm{BASE}}$ in Fig. \ref{fig:observation_new}, fine-tuning the value bias (green rhombus) leads to higher downstream performance than $\boldeq{b}_q$ (red rhombus) and $\boldeq{b}_k$ (blue rhombus). We then analyze the expressiveness of bias terms $\boldeq{b}_q$, $\boldeq{b}_k$, $\boldeq{b}_v$, and provide further evidence supporting our~observation.  

We empirically validate the effectiveness of fine-tuning $\boldeq{b}_v$ versus $\boldeq{b}_q$ and $\boldeq{b}_k$ on BERT$_{\mathrm{BASE}}$ using the GLUE/SuperGLUE benchmarks \cite{wang2018glue,wang2019superglue}, without any post-hoc evaluation. We further extend our validation to autoregressive \glspl{LLM} (such as OPT-1.3B) and bias-free \glspl{LLM} (such as LLaMA2-7B) across various tasks. At the same time, our work is compatible and can be readily combined with \gls{PEFT} methods, e.g., \gls{LoRA} \cite{hu2022lora}, \gls{VeRA} \cite{kopiczko2024vera}, \gls{DoRA} \cite{liu2024dora}, \camera{and PiSSA \cite{meng2024pissa}}. In particular, we show that directly fine-tuning $\boldeq{b}_v$ \camera{using \gls{LoRA} leads to higher downstream performance compared to $\boldeq{b}_q$ and $\boldeq{b}_k$ in Fig. \ref{fig:observation_new}}, with a substantially lower number of parameters.
Our main contributions are:
\vspace{-5pt}
\begin{itemize}
\item We investigate the link between fine-tuning $\boldeq{b}_q$, $\boldeq{b}_k$, and $\boldeq{b}_v$ with the performance of the downstream task, both analytically and empirically. We study and shed light on the expressive power of bias terms $\boldeq{b}_q$, $\boldeq{b}_k$, and $\boldeq{b}_v$ in the query, key, or value projections. 
Our key finding is that \textit{directly fine-tuning $\boldeq{b}_v$ generally leads to higher downstream performance in low-data regimes, in comparison to $\boldeq{b}_q$ and $\boldeq{b}_k$, without requiring any post-hoc evaluation.}

\item We extensively validate the effectiveness of directly fine-tuning $\boldeq{b}_v$ in low-data regimes, without relying on any post-hoc evaluation, across a wide set of \glspl{LLM}, covering both encoder-only and decoder-only architectures with up to 6.7B parameters, including bias-free \glspl{LLM}. Our experiments span a variety of downstream tasks, such as classification, multiple-choice, and generation tasks. Moreover, our results show that fine-tuning $\boldeq{b}_v$ using \gls{LoRA}/\gls{VeRA}/\gls{DoRA} leads to higher downstream performance compared to $\boldeq{b}_q$ and $\boldeq{b}_k$.

\end{itemize}

\section{Bias-Efficient Fine-Tuning (BEFT) for Scaled Dot-Product Attention}

\label{sec:method}
Given an \gls{LLM} model with pre-trained parameters including weights and bias terms, the number of bias terms is substantially smaller than that of weights. In this section, we investigate the link between the bias terms $\boldeq{b}_q$, $\boldeq{b}_k$, and $\boldeq{b}_v$ and the downstream performance. Following the procedure in \cite{zaken2022bitfit}, we first fine-tune all the bias terms and then evaluate the change of different bias terms before and after fine-tuning, i.e., $\Delta \boldeq{b}_{\mathcal{T}}$, where $\boldeq{b}_{\mathcal{T}}$ captures the collection of the single type of bias across Transformer layers $l$ from $1$ to $L$, i.e., $\boldeq{b}_{\mathcal{T}} = \left \{ \boldeq{b}_\mathcal{T}^{(l)}\right \}^L_{l=1}$. In \gls{BEFT}, we aim to identify the bias term, among query $\boldeq{b}_q$, key $\boldeq{b}_k$, and value $\boldeq{b}_v$ for \gls{SDPA}, that has the largest change, denoted as the target bias term, for effective fine-tuning (rather than all bias terms). In other words, the target bias type $\mathcal{T}$ with the largest $\Delta (\boldeq{b}_{\mathcal{T}})$ is:
\begin{align}
\mathcal{T} = \underset{\mathcal{T}\in \{q,\ k,\ v\}}{\arg\max}{\{\Delta (\boldeq{b}_{\mathcal{T}})\}}. \nonumber
\end{align}

\begin{figure}[!t]
    \includegraphics[width=1.0\linewidth]{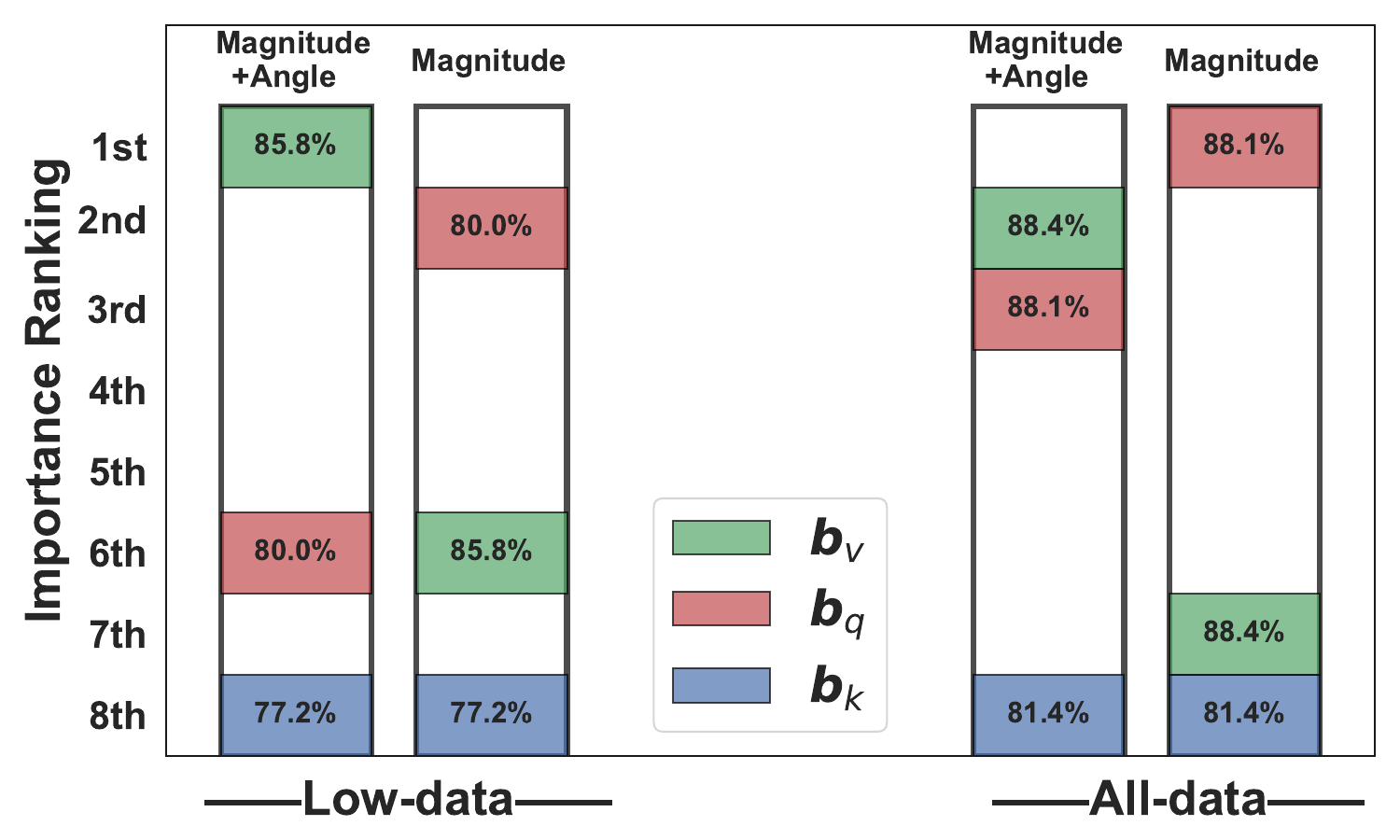}
  \caption{\camera{Importance ranking and accuracy (\%) of fine-tuning different bias terms on SST-2 with low-data regime on BERT$_{\mathrm{BASE}}$. We expect higher-ranked bias terms to achieve higher accuracy. Incorporating angular changes into magnitude changes effectively links the target bias term with the downstream performance.}}
  \label{fig:motivation}
  \vspace{-10pt}
\end{figure}

\begin{figure*}[!t]
     \centering
  \begin{subfigure}{0.3\textwidth}
  \centering
    \includegraphics[width=0.9\linewidth]{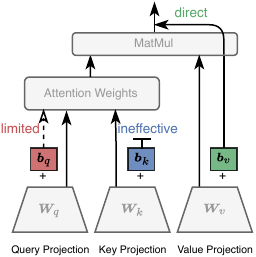}
    \caption{Expressiveness of $\boldeq{b}_q$, $\boldeq{b}_k$, and $\boldeq{b}_v$.}
    \label{fig:main}
  \end{subfigure}
  \begin{subfigure}{0.33\textwidth}
    \includegraphics[width=\linewidth]{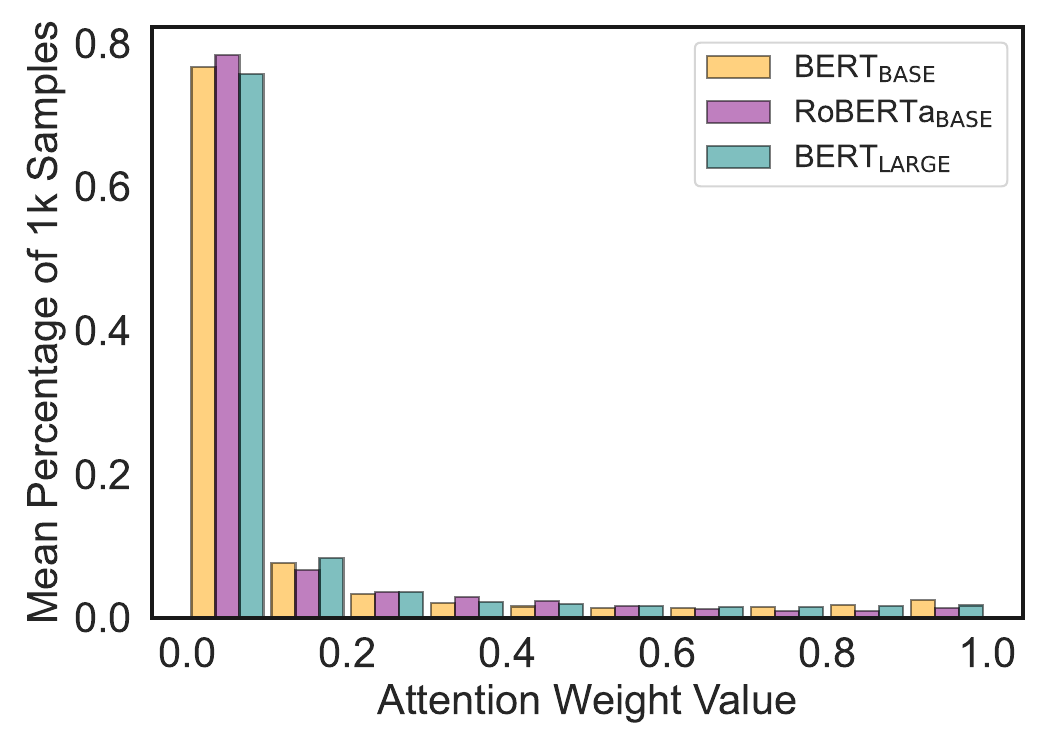}
    \caption{Distribution for Encoder-Only LLMs.}
    \label{fig:main_histo_1}
  \end{subfigure}
  \begin{subfigure}{0.33\textwidth}
    \includegraphics[width=\linewidth]{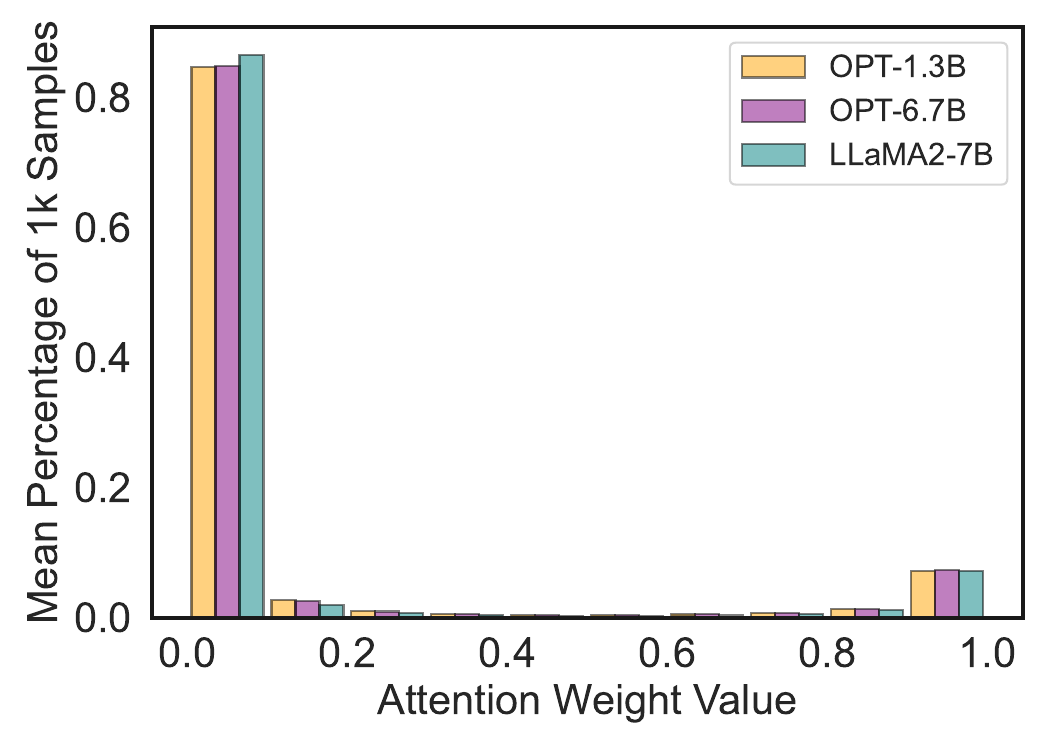}
    \caption{Distribution for Decoder-Only LLMs.}
    \label{fig:main_histo_2}
  \end{subfigure}
    \caption{(a) Fine-tuning the investigated positions—$\boldeq{b}_q$, $\boldeq{b}_k$, and $\boldeq{b}_v$—affects the expressiveness of the low-rank mapping differently: \colorbox{myblue!70}{$\boldeq{b}_k$} is ineffective; \colorbox{myred!70}{$\boldeq{b}_q$} provides only limited improvement; \colorbox{mygreen!70}{$\boldeq{b}_v$} acts effectively as a direct addition after the \gls{SDPA} output. The distributions of attention weights for (b) encoder-only and (c) decoder-only \glspl{LLM} with the mean percentage of $1000$ samples, showing inherent sharpness and high sparsity of attention weights.
    }
\end{figure*}

\camera{In our empirical studies, we observe that the accuracy of fine-tuning $\boldeq{b}_v$ is higher than that of fine-tuning $\boldeq{b}_q$, especially in the low-data regime. In contrast, previous work \cite{zaken2022bitfit} considers only the magnitude of bias change and overall views the change in $\boldeq{b}_q$ is more than $\boldeq{b}_v$, which is inconsistent with our observation considering the downstream performance. To further investigate this and inspired by weight-parameter decomposition 
\cite{liu2024dora,binidecoupling}, we augment the magnitude changes with angular changes to study the link between fine-tuning different bias terms and downstream performance.}

We denote \baichuanchange{the bias term at layer $l$} before fine-tuning as $\boldeq{b}_{\mathcal{T}}^{(l),pre}$ and after fine-tuning as $\boldeq{b}_{\mathcal{T}}^{(l),post}$. We incorporate angular changes into magnitude changes by projection and then scale normalization. Overall, the change of the bias term across all layers is defined as follows (the derivation is in Appendix \ref{sec:appendix_projection_ratio}):
\resizebox{\columnwidth}{!}{$
\begin{aligned}
\Delta(\boldeq{b}_{\mathcal{T}})= 
\frac{1}{L}\sum_{l=1}^{L}\left (1- \frac{\boldeq{b}^{(l),pre}_{\mathcal{T}}\cdot \boldeq{b}^{(l),post}_{\mathcal{T}}}{\max \left( \| \boldeq{b}^{(l),pre}_{\mathcal{T}} \|^2_2,\| \boldeq{b}^{(l),post}_{\mathcal{T}}  \|^2_2\right)} \right ). \nonumber
\end{aligned}$}
\camera{As shown in Fig. \ref{fig:motivation}, augmenting the magnitude with angle effectively links the target bias term with the downstream performance that can be achieved by fine-tuning $\boldeq{b}_v$, compared to $\boldeq{b}_q$ and $\boldeq{b}_k$. Similar patterns have been observed in our empirical evaluation presented in Section \ref{sec:3_2}.}

Next, we discuss the potential expressive power of the bias terms query $\boldeq{b}_q$, key $\boldeq{b}_k$, and value $\boldeq{b}_v$ for \gls{SDPA}. Concurrent with our work, \cite{qiu2025gated} has attracted considerable attention from the research community by showing that gating after the value projection is more effective than after the query or key projection. This study further suggests that gating enhances the expressiveness of low-rank mappings in the attention module. 
Similarly, $\boldeq{b}_{\mathcal{T}}$ in \gls{BEFT} can be interpreted as a linear variant of the additive gating introduced in \cite{qiu2025gated}. As such, $\boldeq{b}_{\mathcal{T}}$ can potentially improve the expressiveness of low-rank mapping.

As shown in Fig. \ref{fig:main}, the query/key/value linear projections are based on $\boldeq{W_{q/k/v}}$ and $\boldeq{b_{q/k/v}}$, as well as the input $\boldeq{X}$. 
For bias-free \glspl{LLM} that tend not to include bias terms in the attention module \cite{touvron2023llama}, we can manually add $\boldeq{b_{q/k/v}}$ in their attention module, denoted as follows: 
\resizebox{\columnwidth}{!}{$
\begin{aligned}
\boldeq{Q}=\boldeq{X}\boldeq{W}_q+\boldeq{b}_q, \boldeq{K}=\boldeq{X}\boldeq{W}_k+\boldeq{b}_k,  \boldeq{V}=\boldeq{X}\boldeq{W}_v+\boldeq{b}_v.\nonumber
\end{aligned}$}
Then, \gls{SDPA} is calculated as follows:
\begin{align}
\mathrm{Attention}(\boldeq{Q},\boldeq{K},\boldeq{V}) = \mathrm{softmax}(\tfrac{\boldeq{Q}\boldeq{K}^T}{\sqrt{d_k}})\boldeq{V}, \nonumber
\end{align}
where $d_k$ is the size of the key vector within one attention head; $\mathrm{softmax}(\tfrac{\boldeq{Q}\boldeq{K}^T}{\sqrt{d_k}})$ is the attention weights $\boldeq{A}$. Given $\boldeq{Q}, \boldeq{K}, \boldeq{V}\in \mathbb{R}^{n\times d_k}$ (n is the sequence length) and softmax as row-wise, we exploit $\boldeq{q}\in \mathbb{R}^{1\times d_k}$ (one row of $\boldeq{Q}$). For $\boldeq{b}_k \in \mathbb{R}^{1\times d_k}$,
\begin{align}
\boldeq{a}=\mathrm{softmax}(\tfrac{\boldeq{q}\boldeq{K}^T}{\sqrt{d_k}}) = \mathrm{softmax}(\tfrac{\boldeq{q}(\boldeq{K}+\boldeq{b}_k)^T}{\sqrt{d_k}}), \nonumber
\end{align}
where $\boldeq{a} \in \mathbb{R}^{1\times n}$ is one row of attention weights $\boldeq{A}$ and $\boldeq{b}_k$ is broadcast across the rows of $\boldeq{K}$, so that each row of $\boldeq{K}$ is incremented by the same elements, while within each row the elements can differ. The constant $\boldeq{q}\boldeq{\boldeq{b}_k}^T \in \mathbb{R}^{1}$ is broadcast across the column of $\boldeq{q}\boldeq{K}^T$. Because softmax is shift-invariant, i.e., $\mathrm{softmax}(\boldeq{q}\boldeq{K}^T) = \mathrm{softmax}(\boldeq{q}\boldeq{K}^T+\boldeq{q}\boldeq{\boldeq{b}_k}^T)$, hence $\boldeq{b}_k $ will not improve expressiveness. 

\begin{figure*}[!ht]
  \centering
  \begin{subfigure}{0.328\textwidth}
    \includegraphics[width=\linewidth]{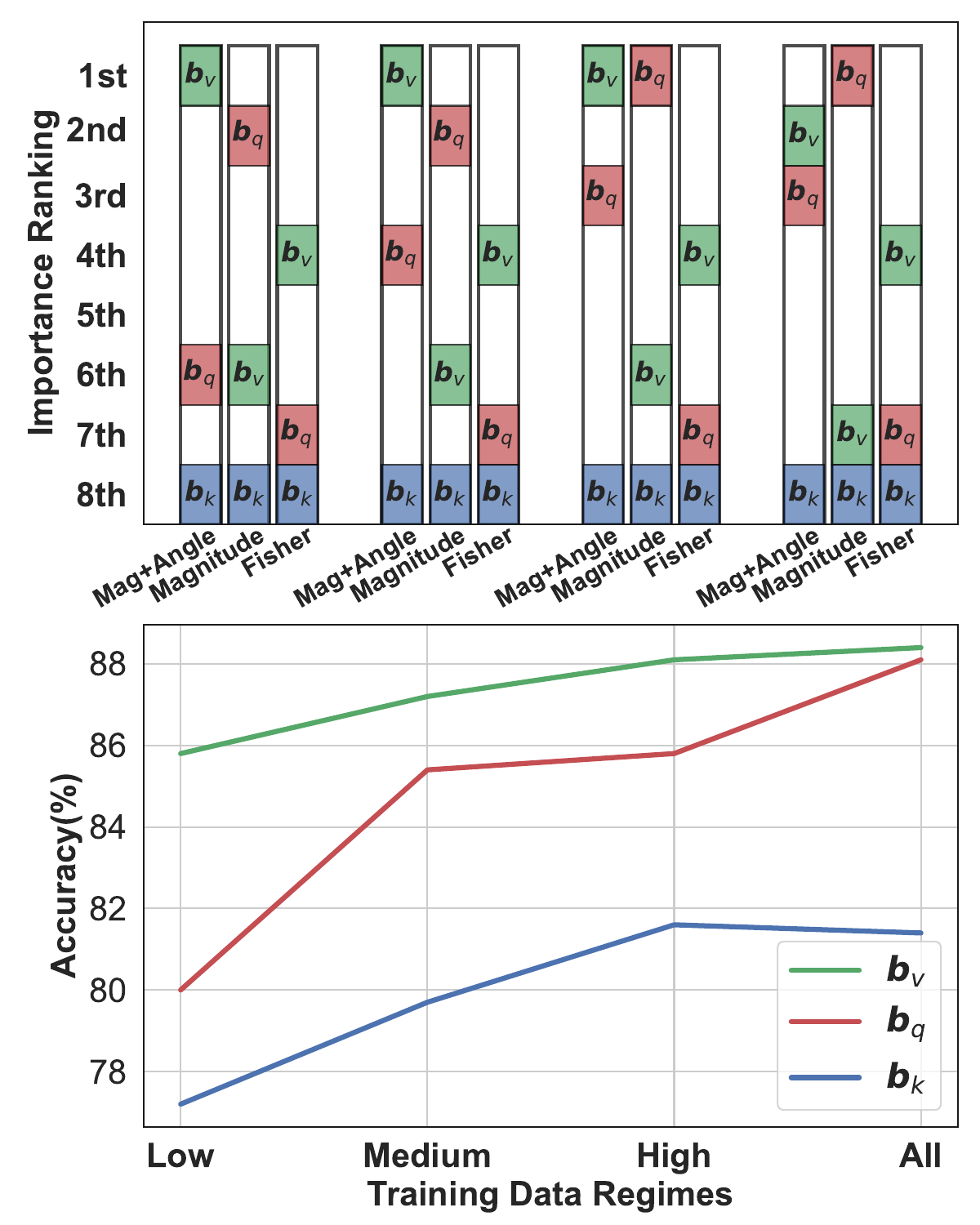}
    \caption{SST-2}
  \end{subfigure}
  \begin{subfigure}{0.328\textwidth}
    \includegraphics[width=\linewidth]{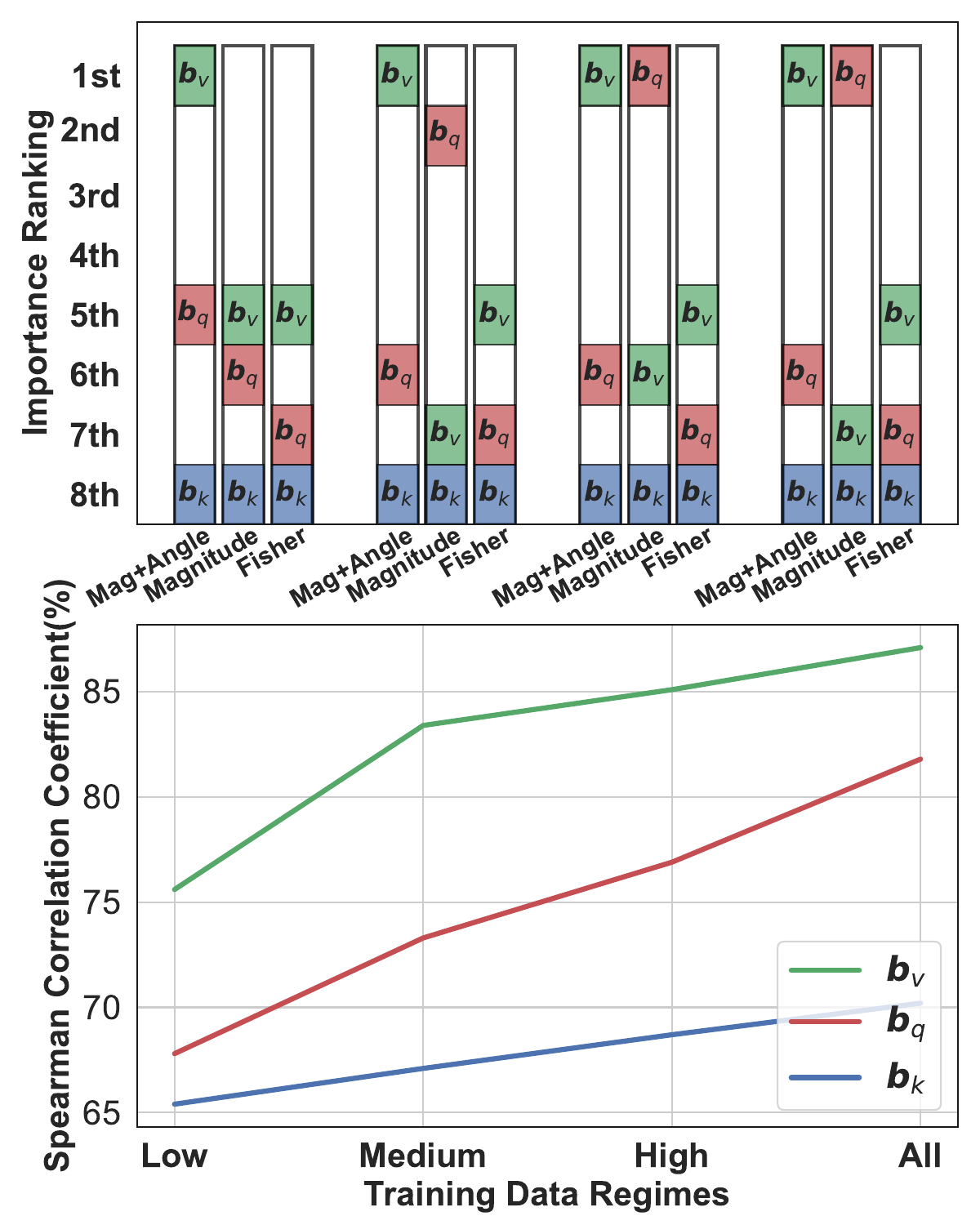}
    \caption{STS-B}
  \end{subfigure}
  \begin{subfigure}{0.328\textwidth}
    \includegraphics[width=\linewidth]{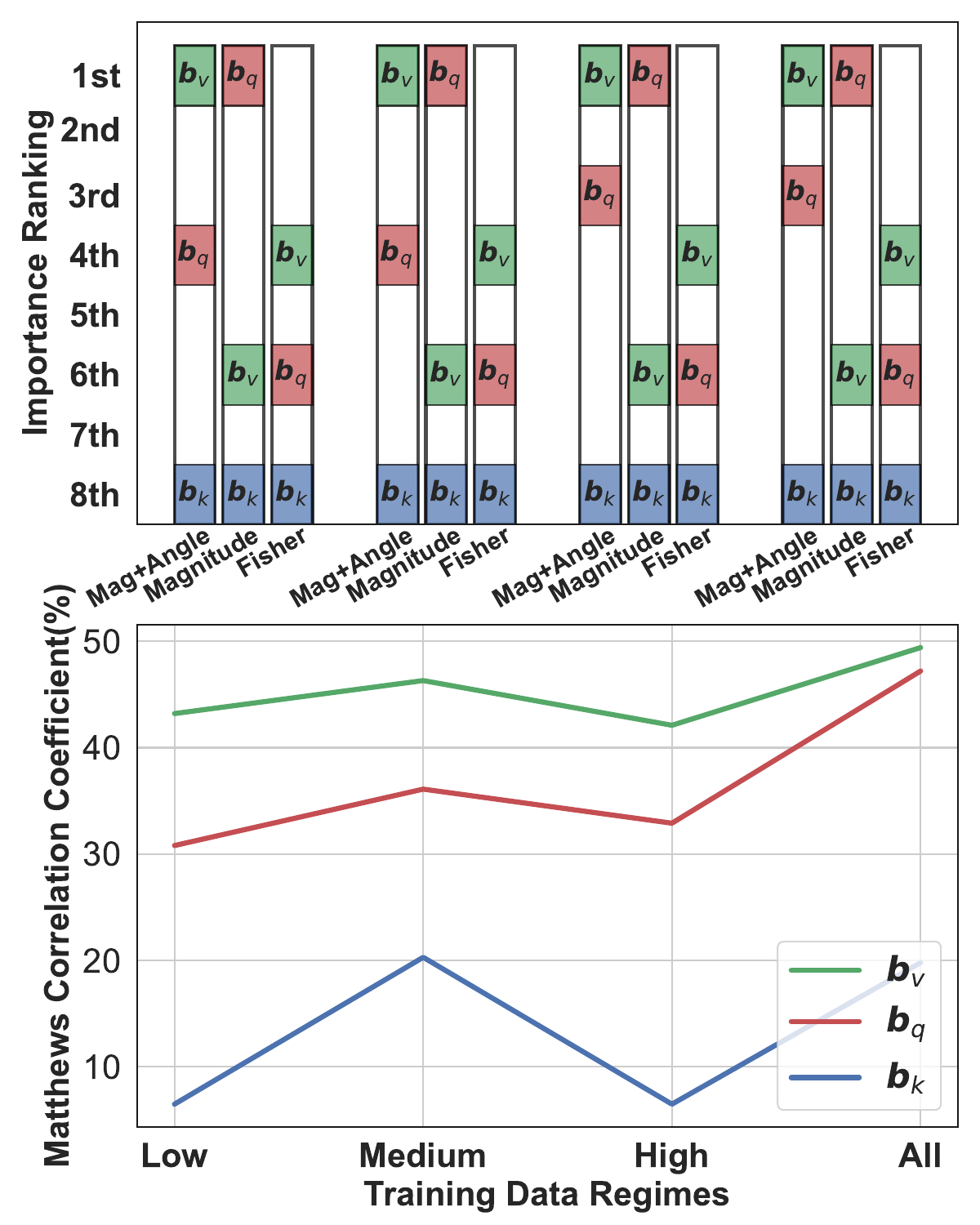}
    \caption{CoLA}
  \end{subfigure}
  \caption{Importance ranking and downstream performance of fine-tuning \colorbox{myred!70}{$\boldeq{b}_q$}, \colorbox{myblue!70}{$\boldeq{b}_k$}, and \colorbox{mygreen!70}{$\boldeq{b}_v$} on (a) SST-2, (b) STS-B, and (c) CoLA, with BERT$_{\mathrm{BASE}}$. Overall, augmenting the magnitude-based approach with angular changes (Mag+Angle) \aclbaichuan{shows a precise and dynamic link between bias-term rankings and downstream performance} across diverse data regimes, outperforming both the Magnitude and Fisher approaches.}
  \label{fig:rank_comparison}
\end{figure*}

For $\boldeq{b}_q \in \mathbb{R}^{1\times d_k}$, $\boldeq{b}_q$ instead affects the attention weights after softmax. The Jacobian of softmax is:
\begin{align}
J = \mathrm{diag}(\boldeq{a})-\boldeq{a}^T\boldeq{a}, \quad |J_{i,j}|  \longrightarrow 0, \forall i,j
\nonumber
\end{align}
where $J \in \mathbb{R}^{n\times n}$ and $\mathrm{diag}$ is the diagonal matrix. The diagonal element is $\boldeq{a}_i(1-\boldeq{a}_i)$ and the off-diagonal element is $-\boldeq{a}_i\boldeq{a}_j$. Moreover, because of $\boldeq{a}_i \in (0,1)$ and $\sum \boldeq{a}_i =1$, the majority of $|J_{i,j}|$ approach zero especially in high-dimension $d_k$, due to the observed inherent sharpness and high sparsity of $\boldeq{A}$ in Fig. \ref{fig:main_histo_1} and \ref{fig:main_histo_2} (as well as Appendix-\ref{sec:appendix_sharpe_sparse}-Fig. \ref{fig:appendix_attention_sink} and \ref{fig:appendix_histogram}). Therefore, softmax is insensitive to the change of $\boldeq{b}_q$ and the capacity of $\boldeq{b}_q$ in improving the expressiveness is limited.

For $\boldeq{b}_v \in \mathbb{R}^{1\times d_k}$, because of $\sum \boldeq{a}_i =1$, we have:
\begin{align}
\boldeq{a}(\boldeq{V}+\boldeq{b}_v)= \boldeq{a}\boldeq{V}+\boldeq{a}(\boldeq{1}_n\boldeq{b}_v)=\boldeq{a}\boldeq{V}+\boldeq{b}_v, \label{eq:bv}
\end{align}
where $\boldeq{b}_v$ is broadcast across the rows of $\boldeq{V}$, i.e., $\boldeq{1}_n\boldeq{b}_v\in \mathbb{R}^{n\times d_k}$, where $\boldeq{1}_n\in \mathbb{R}^{n\times 1}$ denotes an all-ones column vector. As such, $\boldeq{b}_v$ adds a full-dimensional linear freedom in the output space, without being restricted by the softmax or the low-rank attention weights. \aclbaichuan{In this case, $\boldeq{b}_v$ can be regarded as being directly added after the \gls{SDPA} output; gating after \gls{SDPA} and after value projection have been shown to be more effective than other positions \cite{qiu2025gated}. Therefore, $\boldeq{b}_v$ improves the expressiveness more than $\boldeq{b}_q$. We note that in our discussion, we do not include the bias term of the linear layer in output projection, as $\boldeq{b}_v$ is able to yield an effect equivalent to that of bias terms in the output projection (Appendix-\ref{sec:appendix_sdpa_with_output}) in terms of improving the expressiveness.}

In summary, our discussion in this section and empirical evaluation in Section \ref{sec:evaluation} show that \textit{directly fine-tuning $\boldeq{b}_v$ generally leads to higher downstream performance in low-data regimes, in comparison to fine-tuning $\boldeq{b}_q$ or $\boldeq{b}_k$.}\footnote{Note that although the investigation process of bias terms needs post-hoc evaluation, our final conclusion (w.r.t. the efficiency of directly fine-tuning $\boldeq{b}_v$) does not require any post-hoc evaluation.}

\section{Evaluation}\label{sec:evaluation}
\subsection{Experimental Setup}

To validate our approach for effective fine-tuning, we primarily exploit BERT$_{\mathrm{BASE}}$ \cite{devlin2019bert} on the GLUE benchmark \cite{wang2018glue} without the WNLI task \cite{zaken2022bitfit}. Building on this, we extend our key finding to RoBERTa$_{\mathrm{BASE}}$ \cite{liu2019roberta} and BERT$_{\mathrm{LARGE}}$ \cite{devlin2019bert} on the GLUE benchmark and SuperGLUE benchmark \cite{wang2019superglue} (\aclbaichuan{a task-specific final linear classifier layer is used for encoder-only \glspl{LLM}}). Furthermore, to generalize our key finding from masked \glspl{LLM} to autoregressive \glspl{LLM}, we exploit OPT-1.3B and OPT-6.7B \cite{zhang2022opt} evaluated on the GLUE benchmark, SuperGLUE benchmark, SQuAD \cite{rajpurkar2016squad}, and DROP datasets \cite{dua2019drop}, across classification, multiple-choice, and generation tasks. All these experiments undergo training on 1 NVIDIA Tesla T4 \gls{GPU} with 16GB \gls{RAM} for BERT$_{\mathrm{BASE}}$, RoBERTa$_{\mathrm{BASE}}$, and BERT$_{\mathrm{LARGE}}$, with 1 NVIDIA Tesla A40 \gls{GPU} with 48GB \gls{RAM} for OPT-1.3B and 1 NVIDIA Tesla A100 \gls{GPU} with 80GB \gls{RAM} for OPT-6.7B. The experimental details are documented in Appendix \ref{sec:appendix_hyperparameters}. \camera{We also provide the paired t-test for main results to show the statistical significance in Appendix-\ref{sec:appendix_significance}.}

\begin{table*}[!ht]
\centering
\resizebox{\textwidth}{!}{
\begin{tabular}{cccccccccccc}
\toprule[2pt]
Training Data& $\boldeq{b}_\mathcal{T}$ & SST-2  & RTE & QQP & QNLI & MNLI$_m$ & MNLI$_{mm}$ & CoLA & MRPC & STS-B & Avg. \\
\midrule[1pt]
\rowcolor{mygreen!40} 
\cellcolor{white} \multirow{3}{*}{Low}   & 
 $\boldeq{b}_v$  &  \textbf{85.8}$^{\dagger}$   &  \textbf{59.5}$^{\dagger}$ &\textbf{68.8}$^{\dagger}$ & \textbf{73.8}$^{\dagger}$& \textbf{43.8}$^{\dagger}$    & \textbf{45.7}$^{\dagger}$  &\textbf{43.2}$^{\dagger}$&\textbf{84.0}$^{\dagger}$& \textbf{75.6}$^{\dagger}$ &   \textbf{64.5} \\
   &   $\boldeq{b}_q$ &  80.0   &   46.9    &65.1 & 67.7 & 40.1  & 40.5  &30.8 &81.1 &67.8 & 57.8  \\
   & $\boldeq{b}_k$  &  77.2   & 46.5      &60.5& 66.9&  39.5   &40.1   &6.5&79.2&65.4& 53.5   \\
\hline
\rowcolor{mygreen!40}
\cellcolor{white} \multirow{3}{*}{Medium}     &  $\boldeq{b}_v$ &  \textbf{87.2}$^{\dagger}$   &  \textbf{65.7}$^{\dagger}$    &\textbf{73.5}$^{\dagger}$&\textbf{79.3}$^{\dagger}$ &  \textbf{58.3}$^{\dagger}$  & \textbf{59.8}$^{\dagger}$  &\textbf{46.3}$^{\dagger}$&\textbf{84.1}$^{\dagger}$&\textbf{83.4}$^{\dagger}$& \textbf{70.8}   \\ 
   & $\boldeq{b}_q$  &   85.4    &  58.1      &69.3  & 72.9  &  45.0   & 46.4   &36.1 &83.1 &73.3 &   63.3 \\
    & $\boldeq{b}_k$  & 79.7    & 54.5      &0.00&67.6 &  41.4   & 42.1  &20.3&80.8&67.1&  50.4  \\
\hline
\rowcolor{mygreen!40}
\cellcolor{white}
\multirow{3}{*}{High}      &  $\boldeq{b}_v$ &   \textbf{88.1}$^{\dagger}$  &   \textbf{62.8}$^{\dagger}$    &\textbf{74.4}$^{\dagger}$& \textbf{80.8}$^{\dagger}$&  \textbf{62.2}$^{\dagger}$   & \textbf{64.2}$^{\dagger}$  &\textbf{42.1}$^{\dagger}$&\textbf{85.5}$^{\dagger}$&\textbf{85.1}$^{\dagger}$&  \textbf{71.7}  \\
   &  $\boldeq{b}_q$ &   85.8   &  58.4      &71.8  & 76.6   & 47.8   & 48.5   &32.9 &84.2 &76.9 &  64.7  \\
 & $\boldeq{b}_k$  &  81.6  & 53.0      &12.9& 68.1&  42.1   & 42.1  &6.5&78.8&68.7&   50.4 \\
\bottomrule[2pt]
\end{tabular}}
\caption{Downstream performance (\%) of fine-tuning \baichuanchange{different bias terms} with BERT$_{\mathrm{BASE}}$ on the GLUE benchmark. We highlight the \textbf{best} results and annotate the \baichuanchange{target} bias term selected by our approach with the symbol ${\dagger}$. \aclbaichuan{For the datasets shown in Fig. \ref{fig:rank_comparison}, we additionally report the mean$\pm$std over three random seeds in Appendix \ref{sec:appendix_multi-seed}-Table \ref{tab:appendix_multi-seed_table_1}.}
}
\label{tab:main}
\end{table*}

\subsection{Bias-Term in Value vs Query and Key}

\label{sec:3_2}
We report the importance ranking and downstream performance for various bias-selection approaches, as shown in Fig. \ref{fig:rank_comparison} and Table \ref{tab:main}, \aclbaichuan{where the importance ranking is obtained by sorting the change of bias terms, as discussed in Section \ref{sec:method}. The greater the change, the higher the importance ranking}.

In Fig. \ref{fig:rank_comparison}, we present results on three representative datasets from the GLUE benchmark with distinct evaluation indicators, and we reveal the interpretative finding across different data regimes: low-, medium-, high-, and all-data regimes.

In the SST-2 dataset, as shown in Fig. \ref{fig:rank_comparison} (a), our approach consistently selects $\boldeq{b}_v$ as the \baichuanchange{target} bias, while the Magnitude approach always selects $\boldeq{b}_q$. However, the accuracy corresponding to fine-tuning $\boldeq{b}_v$ surpasses that of $\boldeq{b}_q$ across different data regimes, showing the inaccuracy of the Magnitude approach—particularly in the low-data regime. In the case of the Fisher approach, although it also identifies $\boldeq{b}_v$ as the \baichuanchange{target} bias, it \baichuanchange{results in a static importance ranking for bias terms across different data regimes}. In contrast, our approach dynamically captures the narrowing importance gap between $\boldeq{b}_v$ and $\boldeq{b}_q$ as training data increases, aligning closely with the observed convergence in their downstream performance. These results demonstrate that our approach not only achieves superior identification of \baichuanchange{target} bias than the Magnitude approach, but also has the capacity of dynamic importance ranking compared to the Fisher approach (\baichuanchange{see Appendix \ref{sec:appendix_qualitative_observations}-Fig. \ref{fig:appendix_rank_comparison} for other datasets}).

Similarly, for the STS-B dataset, as shown in Fig. \ref{fig:rank_comparison} (c), the Magnitude approach selects $\boldeq{b}_q$ as the \baichuanchange{target} bias from medium- to all-data regimes. However, selecting $\boldeq{b}_q$ has a lower performance (i.e., lower Spearman correlation coefficient) than selecting $\boldeq{b}_v$. Although the Magnitude approach selects $\boldeq{b}_v$ in the low-data regime, the importance ranking gap between $\boldeq{b}_v$ and $\boldeq{b}_q$ does not align with the observed differences in downstream performance. %, indicating a lack of adaptability. 
The Fisher approach, on the other hand, suffers from \baichuanchange{the static importance ranking} as before. Conversely, from low- to medium-data regimes, our approach reflects the increasing performance gap between $\boldeq{b}_v$ and $\boldeq{b}_q$, in line with their changing importance ranking. 

Furthermore, for the CoLA dataset, as shown in Fig. \ref{fig:rank_comparison} (b),  our bias-efficient approach consistently selects $\boldeq{b}_v$ as the \baichuanchange{target} bias, while the Magnitude approach always selects $\boldeq{b}_q$. Considering the performance metric (i.e., the Matthews correlation coefficient), the Magnitude approach selects the incorrect \baichuanchange{target} bias between $\boldeq{b}_v$ and $\boldeq{b}_q$. Moreover, the Magnitude approach exhibits a static importance ranking across different data regimes. Similarly, the Fisher approach results in a \baichuanchange{fixed importance ranking}. In contrast, our approach adaptively adjusts the performance gap between $\boldeq{b}_v$ and $\boldeq{b}_q$, aligning with the observed shift in their importance rankings. 
These results confirm the advantages of augmenting the magnitude with angular changes over Magnitude and Fisher approaches.

In Table \ref{tab:main}, we report the downstream performance for $\boldeq{b}_q$, $\boldeq{b}_k$, and $\boldeq{b}_v$. We annotate the \baichuanchange{target} bias term selected by our approach with the symbol ${\dagger}$. In Table \ref{tab:main}, fine-tuning $\boldeq{b}_v$ consistently achieves higher performance than fine-tuning $\boldeq{b}_q$ across the GLUE benchmark, from low- to high-data regimes. Conversely, fine-tuning $\boldeq{b}_k$ exhibits consistently lower performance than fine-tuning $\boldeq{b}_q$. As such, our approach consistently succeeds across all datasets and diverse training data regimes. These results are also consistent with the discussion in Section \ref{sec:method}. Overall, these results show that our approach captures importance rankings and selects the \baichuanchange{target} bias term to be effectively fine-tuned.

\subsection{Efficiency and Effectiveness}

\baichuanchange{Prior work has indicated that fine-tuning all-bias-terms has competitive performance to full-parameter fine-tuning in low-data regimes \cite{zaken2022bitfit,logan2022cutting,doering2024empirical}. Accordingly, we investigate the efficiency and effectiveness of our \gls{BEFT} in low-data regimes.}

\begin{table}[!t]
\centering
\resizebox{\columnwidth}{!}{
\begin{tabular}{ccccc}
\toprule[2pt]
Fine-Tuning & Params$\downarrow$    & Runtime$\downarrow$ & Accuracy$\uparrow$ \\
\midrule[1pt]
\rowcolor{mygreen!40}
Our \gls{BEFT} & \textbf{0.01\%} & \textbf{132.9} &  
58.53\text{\scriptsize$\pm$1.88}\\
Rand uniform  & 0.01\% & 659.6 &  50.40\text{\scriptsize$\pm$2.94} \\
%50.2 \\
All biases & \underline{0.09\%}  & \underline{144.9} &  
56.40\text{\scriptsize$\pm$2.88}\\
All parameters & 100\%  & 206.1 & 
57.46\text{\scriptsize$\pm$2.20}\\
\bottomrule[2pt]
\end{tabular}}
\caption{\baichuanchange{Runtime (s) and downstream performance (\%) (\aclbaichuan{mean$\pm$std over three runs with different random seeds}). We also highlight the \underline{second-best} results, indicating the strong potential of fine-tuning the target bias term.}}
\label{tab:main_side}
\end{table}

In Table~\ref{tab:main_side}, our \gls{BEFT} achieves remarkable parameter efficiency with only 0.01\% of the full parameters, when fine-tuning BERT$_{\mathrm{BASE}}$ in the low-data regime of the RTE dataset. Fine-tuning all biases \cite{zaken2022bitfit} requires 0.09\% of the full parameters, nearly 9$\times$ more trainable parameters. \baichuanchange{Although ``rand uniform'' samples and fine-tunes the same number of bias terms as \gls{BEFT}, it yields substantially worse performance and requires more time due to inefficient uniform sampling.} In terms of runtime, full-parameter fine-tuning takes 206.1 \gls{seconds}, and all-bias fine-tuning requires 144.9 \gls{seconds}, whereas \gls{BEFT} completes in just 132.9 \gls{seconds}. Despite this significant reduction in parameter footprint and a decrease in training time, \baichuanchange{our proposed \gls{BEFT} achieves a comparable performance with all-bias fine-tuning and all-parameter fine-tuning in this case (more cases in Appendix \ref{sec:appendix_qualitative_observations}-Fig. \ref{fig:comparison_rte}).} These results highlight \gls{BEFT}'s promise as an unprecedented parameter-efficient fine-tuning strategy.

\subsection{Extension to Other Datasets and \glspl{LLM}}
\label{sec:result_extensive}
Motivated by the preceding findings, it is natural to ask: Can the key finding (\textit{w.r.t. the efficiency of directly fine-tuning $\boldeq{b}_v$}) be extended to different datasets and \glspl{LLM}, \aclbaichuan{and combination with \gls{PEFT} methods (such as \gls{LoRA} \cite{hu2022lora}, \gls{VeRA} \cite{kopiczko2024vera}, and \gls{DoRA} \cite{liu2024dora})}, \textit{without requiring any post-hoc evaluation}? {The short answer is yes}. 

\begin{table}[!t]
\centering
\resizebox{\columnwidth}{!}{
\begin{tabular}{cccccc}
\toprule[2pt]
 \multirow{2}{*}{Models}  & \multirow{2}{*}{$\boldeq{b}_\mathcal{T}$} & \multicolumn{2}{c}{GLUE}  & \multicolumn{2}{c}{SuperGLUE}\\
&  & SST-2  & RTE  &  CB & WiC\\
\midrule[1pt]
\rowcolor{mygreen!40}
\cellcolor{white}
\multirow{3}{*}{BERT$_{\mathrm{BASE}}$} & $\boldeq{b}_v$ & \textbf{85.8} & \textbf{59.5}   & \textbf{59.0} & \textbf{69.6} \\
 & $\boldeq{b}_q$ & 80.0  & 46.9   &  47.2 & 66.1\\
& $\boldeq{b}_k$ & 77.2 & 46.5  & 43.3 & 62.8 \\
\hline
\rowcolor{mygreen!40}
\cellcolor{white}
\multirow{3}{*}{RoBERTa$_{\mathrm{BASE}}$} & $\boldeq{b}_v$ & \textbf{88.6} &  \textbf{56.7} & \textbf{90.5} & \textbf{62.1} \\
 & $\boldeq{b}_q$ & 87.0 & 56.3  & 85.0 &57.8 \\
& $\boldeq{b}_k$ & 72.0 & 52.3  & 76.2  & 57.9 \\
\hline
\rowcolor{mygreen!40}
\cellcolor{white}
\multirow{3}{*}{BERT$_{\mathrm{LARGE}}$} & $\boldeq{b}_v$ & \textbf{89.2} & \textbf{60.6}  &  \textbf{58.7} & \textbf{66.6} \\
& $\boldeq{b}_q$ & 83.8 &  53.4  & 50.9 &61.2 \\
& $\boldeq{b}_k$ & 69.3 & 53.4  & 26.6 & 61.2\\
\bottomrule[2pt]
\end{tabular}}
\caption{The key finding (\textit{directly fine-tuning $\boldeq{b}_v$}) can be extended to different datasets and \glspl{LLM} without requiring any post-hoc evaluation. \aclbaichuan{We additionally report the mean$\pm$std over multi-seeds in Appendix \ref{sec:appendix_multi-seed}-Table \ref{tab:appendix_multi-seed_table_3}.}} 
\label{tab:other_models}
\end{table}

\paragraph{Extension to different datasets:} To evaluate the extension of our key finding to different datasets in the low-data regime, we exploit the CB and WiC datasets from the SuperGLUE benchmark \cite{wang2019superglue}, where the CB dataset contains only 250 training examples. First, we evaluate the same model, i.e., BERT$_{\mathrm{BASE}}$, on these different datasets. As presented in Table \ref{tab:other_models}, for BERT$_{\mathrm{BASE}}$, fine-tuning $\boldeq{b}_v$ is better than fine-tuning $\boldeq{b}_q$; fine-tuning $\boldeq{b}_q$ is better than fine-tuning $\boldeq{b}_k$. \textcolor{black}{We also extend our \gls{BEFT} to multilingual and commonsense reasoning datasets in Appendix-\ref{sec:appendix_more_data}}. These results consistently mirror our preceding finding on the GLUE benchmark, showing that \textit{directly fine-tuning $\boldeq{b}_v$} can generalize to new datasets.

\paragraph{Extension to different \glspl{LLM}:} To evaluate the extension of our key finding to different \glspl{LLM} in the low-data regime, we exploit the RoBERTa$_{\mathrm{BASE}}$ \cite{liu2019roberta} and BERT$_{\mathrm{LARGE}}$ \cite{devlin2019bert}. First, on SST-2 and RTE, fine-tuning $\boldeq{b}_v$ consistently outperforms fine-tuning $\boldeq{b}_q$, as presented in Table \ref{tab:other_models}. This trend holds when extending RoBERTa$_{\mathrm{BASE}}$ and BERT$_{\mathrm{LARGE}}$ to CB and WiC.

\begin{table}[!t]
\centering
\resizebox{\columnwidth}{!}{
\begin{tabular}{cccccc}
\toprule[2pt]
$\boldeq{b}_\mathcal{T}$  &Directly &+\gls{LoRA}  & +\gls{VeRA} & +\gls{DoRA} \\ 
\midrule[1pt]
$\boldeq{b}_v$ & \cellcolor{mygreen!40}\textbf{85.8} &\cellcolor{mygreen!40}\textbf{85.0}  &\cellcolor{mygreen!40}\textbf{82.2}  & \cellcolor{mygreen!40}\textbf{86.0} \\
$\boldeq{b}_q$ & 80.0 & 82.6 &81.8 & 83.5 \\
$\boldeq{b}_k$ & 77.2 & 76.9 &81.7 & 76.9 \\
\hline
$\Delta$Params$\downarrow$& 100\% & 58.3\% &4.2\%  & 59.4\% \\
\bottomrule[2pt]
\end{tabular}}
\caption{\aclbaichuan{The downstream performance (\%) of variants of \gls{LoRA}/\gls{VeRA}/\gls{DoRA} designed for \gls{BEFT} on SST-2 with BERT$_{\mathrm{BASE}}$ , where $\boldeq{b}_v$ surpasses $\boldeq{b}_q$ and $\boldeq{b}_k$.}}
\label{tab:vera_beft}
\end{table}

\begin{table*}[!ht]
\centering
\resizebox{\textwidth}{!}{
\begin{tabular}{cccccccccccccc}
\toprule[2pt]
 \multirow{2}{*}{LLMs}   & Fine-Tuning & \multirow{2}{*}{Params}  & SST-2  & RTE & CB &BoolQ &WSC& WiC  & MultiRC & COPA & ReCoRD & SQuAD & DROP  \\
 &Techniques&& \multicolumn{7}{c}{--------------------classification--------------------}& \multicolumn{2}{c}{--multiple choice--}&\multicolumn{2}{c}{---generation---}\\
\midrule[1pt]
\rowcolor{mygreen!40}
\cellcolor{white}
\multirow{7}{*}{OPT-1.3B} 
&  $\boldeq{b}_v$ & 0.04\textperthousand& \underline{93.1}$^{\dagger}$ & \underline{71.5}$^{\dagger}$ & \textbf{80.3}&\underline{65.8}&\textbf{63.5}&\textbf{61.6}&\underline{69.0}&\textbf{77.0}&\textbf{71.5}&\textbf{79.8}&\textbf{28.2} \\
 & $\boldeq{b}_q$ &0.04\textperthousand& 72.7 & 55.6&\underline{71.4}&60.4&\underline{62.5}&\underline{59.2}&56.3&74.0&\underline{71.2}&69.3&21.4  \\
 &  $\boldeq{b}_k$ &0.04\textperthousand& 53.5 &  53.1&37.5&45.3&44.2&57.0&45.3&75.0&70.5&24.2&10.4 \\
 &LoRA  & 1.2\textperthousand& \aclbaichuan{\textbf{93.6}} & \aclbaichuan{\textbf{73.2}}& \aclbaichuan{69.6}&\aclbaichuan{\textbf{74.3}}&\textbf{63.5}&50.0&\aclbaichuan{\textbf{73.2}}&\underline{76.0}&71.0&\underline{79.7}&\underline{27.9} \\
 &Prefix &0.4\textperthousand & 91.9& 63.9&\textbf{80.3}& 64.3&61.5&57.6&60.0&69.0&70.5&78.3&25.2\\
 &ICL &0\textperthousand&80.3&53.0&48.2&58.5&47.1&50.6&46.3&69.0&70.9&58.6&20.3\\
 &Zero-Shot &0\textperthousand& 53.5&53.0&39.3&45.5&44.2&57.3&45.3&75.0&70.5&27.2&11.1\\
 \hline
 \rowcolor{mygreen!40}
\cellcolor{white}
 \multirow{7}{*}{OPT-6.7B} 
&  $\boldeq{b}_v$ & 0.02\textperthousand & \textbf{95.2} & \textbf{82.6}& \textbf{96.4}&\underline{78.8}&\textbf{63.5}&\textbf{65.2}&\underline{73.9}&\textbf{83.0}&\textbf{77.7}&\textbf{86.9}& \textbf{32.4}\\
 &$\boldeq{b}_q$ &0.02\textperthousand & 80.3&62.4&67.8&64.7&\underline{60.5}&\underline{59.4}&58.6&\textbf{83.0}&76.5&75.5&27.9\\
 &$\boldeq{b}_k$ &0.02\textperthousand&61.2&54.8&
 50.0&59.5&37.5&51.2&44.5&82.0&76.0&37.4&16.8\\
 &LoRA & 0.6\textperthousand&\aclbaichuan{\underline{95.1}}&\aclbaichuan{\underline{81.9}}&\aclbaichuan{75.0}&\aclbaichuan{\textbf{79.1}}&\textbf{63.5}&50.4&\aclbaichuan{\textbf{74.4}}&82.0&76.3&\underline{85.7}&\underline{31.1}\\
 &Prefix & 0.2\textperthousand &\underline{95.1}&78.3&\underline{89.2}&72.6&59.6&54.7&58.9&78.0&\underline{76.8}&84.6&30.5\\
 &ICL & 0\textperthousand& 84.6&65.7&57.1&68.9&50.9&53.6&50.4&82.0&\underline{76.8}&74.2&27.5\\
 &Zero-Shot &0\textperthousand&61.2&55.2&51.7&59.5&37.5&51.2&44.5&82.0&76.1&36.5&17.7\\
\bottomrule[2pt]
\end{tabular}}
\caption{Downstream performance (\%) of fine-tuning techniques in low-data regimes. \aclbaichuan{For our key finding, we additionally report the mean$\pm$std over three random seeds in Appendix \ref{sec:appendix_multi-seed}-Table \ref{tab:appendix_multi-seed_table_5}.} \baichuanchange{\gls{BEFT} demonstrates competitive performance with mainstream \gls{PEFT} techniques regarding parameter-efficiency and downstream~adaptation}.}
\label{tab:opt}
\end{table*}

\setlength{\tabcolsep}{6pt}

\aclbaichuan{
\paragraph{Extend to \gls{LoRA}/\gls{VeRA}/\gls{DoRA} for Bias Terms:} To further evaluate the extension of our key finding, we inject \gls{LoRA} \cite{hu2022lora}, \gls{VeRA} \cite{kopiczko2024vera}, and \gls{DoRA} \cite{liu2024dora} into bias terms (please see Appendix-\ref{sec:appendix_lora_vera_dora}), investigating the capacity of \gls{BEFT} with these \gls{PEFT} in low-data regimes \cite{schulman2025lora}. }

\aclbaichuan{
As presented in Table \ref{tab:vera_beft}, with $(q,k,r) = (24,32,8)$ for BERT$_{\mathrm{BASE}}$, by injecting \gls{LoRA} or \gls{DoRA} into the bias terms, more than 40\% of the trainable parameters can be reduced, while over 95\% for \gls{VeRA}. At the same time, the downstream performance of fine-tuning $\boldeq{b}_v$ with these \gls{PEFT} methods consistently outperforms fine-tuning $\boldeq{b}_q$ and $\boldeq{b}_k$. The key finding can also be directly combined with \gls{LoRA}/\gls{VeRA}/\gls{DoRA} (more variants and \glspl{LLM} in Appendix-\ref{sec:appendix_lora_vera_dora} Table \ref{tab:vera_1d} and \ref{tab:lora_beft}).} \textcolor{black}{Moreover, we also extend our \gls{BEFT} to more advanced \gls{PEFT}, such as PiSSA \cite{meng2024pissa} and BA-LoRA \cite{chang2026balora}, in Appendix-\ref{sec:appendix_pissa}}.

Overall, these results show that \textit{directly fine-tuning $\boldeq{b}_v$} can be effective to different datasets and \glspl{LLM}, and be combined with \gls{LoRA}, \gls{VeRA}, and \gls{DoRA}, without requiring any post-hoc~evaluation. 

\subsection{Extension to Autoregressive \glspl{LLM}}

To further validate the generality of our finding from masked \glspl{LLM} to autoregressive \glspl{LLM}, we consider OPT-1.3B and OPT-6.7B \cite{zhang2022opt} on various benchmarks, including the GLUE \cite{wang2018glue}, SuperGLUE \cite{wang2019superglue}, SQuAD \cite{rajpurkar2016squad}, and DROP datasets \cite{dua2019drop} covering classification, multiple-choice, and generation tasks. We also compare our \gls{BEFT} against various mainstream \gls{PEFT} methods, such as LoRA \cite{hu2022lora}, prefix tuning \cite{li2021prefix}, \gls{ICL} \cite{brown2020language} and zero-shot techniques \cite{brown2020language}, as presented in Table \ref{tab:opt}. First, fine-tuning $\boldeq{b}_v$ consistently performs the best among $\boldeq{b}_v$, $\boldeq{b}_q$, and $\boldeq{b}_k$, which demonstrates the effective generalization of our key finding (for instance, the results of SST-2 with OPT-1.3B match the target bias in Appendix \ref{sec:appendix_qualitative_observations}-Fig. \ref{fig:opt_target_bias}). Second, our \gls{BEFT} requires 30$\times$ and 10$\times$ fewer parameters than \gls{LoRA} and prefix tuning, respectively, while maintaining competitive or superior performance across tasks. \aclbaichuan{In addition, our \gls{BEFT} also demonstrates the competence when compared to more advanced techniques such as \gls{VeRA} \cite{kopiczko2024vera} and \gls{DoRA} \cite{liu2024dora} in Appendix-\ref{sec:appendix_sota_peft}}. Finally, while \gls{ICL} and zero-shot techniques do not require fine-tuning, they exhibit inherently inferior performance compared to our \gls{BEFT}. Overall, these results indicate that our \gls{BEFT} approach achieves competitive performance and efficiency relative to the mainstream \aclbaichuan{and more advanced} \gls{PEFT} techniques.

\begin{table}[!t]
\centering
\resizebox{\columnwidth}{!}{
\begin{tabular}{ccccc}
\toprule[2pt]
 LLaMA2-7B  & Params & SST-2 & COPA & SQuAD\\
\midrule[1pt]
\rowcolor{mygreen!40}
Adding $\boldeq{b}_v$ & 0.02\textperthousand & \textbf{94.9}\text{\scriptsize$\pm$0.7}  & \textbf{85.0}\text{\scriptsize$\pm$0.0} & \textbf{89.4}\text{\scriptsize$\pm$0.8}\\
Adding $\boldeq{b}_q$ &0.02\textperthousand &90.0\text{\scriptsize$\pm$0.3}  & 79.0\text{\scriptsize$\pm$0.0} & 88.0\text{\scriptsize$\pm$0.6}\\
Adding $\boldeq{b}_k$ &0.02\textperthousand &68.0\text{\scriptsize$\pm$0.4}& 79.0\text{\scriptsize$\pm$0.0} & 82.0\text{\scriptsize$\pm$0.5}\\
\hline
LoRA & 0.6\textperthousand & 95.6\text{\scriptsize$\pm$0.1}& 83.3\text{\scriptsize$\pm$0.5}&90.4\text{\scriptsize$\pm$0.6} \\
Prefix & 0.2\textperthousand & 93.5\text{\scriptsize$\pm$0.4}& 79.0\text{\scriptsize$\pm$0.0}&89.9\text{\scriptsize$\pm$1.8} \\
\bottomrule[2pt]
\end{tabular}}
\caption{Downstream performance (\%) (\aclbaichuan{mean$\pm$std over three runs with different random seeds}) of adding bias into bias-free \glspl{LLM}, where our key finding still~holds.}
\label{tab:llama_beft}
\end{table}

\aclbaichuan{
\subsection{Extension to Bias-Free \glspl{LLM}}
Today, certain \glspl{LLM} are bias-free, consequently omitting bias terms in their attention modules \cite{touvron2023llama,guo2024deepseek,team2024gemma}. Nevertheless, methods such as per-layer bias-terms injection \cite{li2023inference}, scaling and biasing operation \cite{wu2024advancing,wu2024reft}, and activating only layer-wise additive biases \cite{sinii-etal-2025-steering} have shown both effectiveness and efficiency. To validate our key finding with bias-free \glspl{LLM}, we manually add different bias terms to their attention module. 
As shown in Table \ref{tab:llama_beft}, we add $\boldeq{b}_v$, $\boldeq{b}_q$, and $\boldeq{b}_k$ into LLaMA (LLaMA-2-7B-hf) \cite{touvron2023llama}. The downstream performance of fine-tuning additive $\boldeq{b}_v$ still surpasses that of fine-tuning additive $\boldeq{b}_q$ and additive $\boldeq{b}_k$. Similar patterns are observed and presented for GPT-J-6B \cite{gpt-j} and DeepSeek-Coder-Base-1.3B \cite{guo2024deepseek} in Appendix-\ref{sec:appendix_more_llm_beft}. These results show that for bias-free \glspl{LLM} in the low-data regime, adding $\boldeq{b}_v$ and fine-tuning $\boldeq{b}_v$ can lead to competitive performance, without requiring any post-hoc evaluation.} In addition, adding $\boldeq{b}_v$ requires 30$\times$ and 10$\times$ fewer parameters than \gls{LoRA} and prefix tuning, respectively, while maintaining competitive performance across tasks.

\section{Related Work}

\subsection{Parameter-Efficient Fine-Tuning (PEFT)}

\gls{PEFT} techniques aim to fine-tune only a small subset of parameters in \glspl{LLM}, thereby significantly reducing resource overheads. Addition-based \gls{PEFT} methods introduce new trainable components into the model architectures. Prefix tuning \cite{li2021prefix} adds trainable prefix tokens to attention blocks; adapter tuning \cite{houlsby2019parameter} inserts a trainable adapter module into Transformer layers; and prompt-tuning \cite{lester2021power} \baichuanchange{introduces a sequence of trainable embeddings prepended to the input}. Reparameterization-based \gls{PEFT} methods change the representation of the parameter or its update. LoRA \cite{hu2022lora} exploits the low-rank decomposition to approximate the original parameter change, and GaLore \cite{zhao2024galore} reparameterizes the gradient into low-rank matrices. While effective, these \gls{PEFT} techniques \baichuanchange{are limited by their out-of-the-box usability, with extra requirements for additional configuration and
auxiliary reparameterization.}

\subsection{Bias-Efficient Fine-Tuning}

Bias-efficient fine-tuning techniques aim to fine-tune only the bias terms in \glspl{LLM}. Fine-tuning all biases—as in BitFit \cite{zaken2022bitfit}—has been shown to achieve competitive performance with full-parameter fine-tuning, especially in low-data regimes \cite{zaken2022bitfit,logan2022cutting}. This advantage is attributed in part to the low intrinsic dimensionality of pre-trained \glspl{LLM}, which allows meaningful adaptation within a restricted subspace \cite{aghajanyan2021intrinsic,mosbachstability,tang2024amu}. Prior papers reveal that different bias terms contribute differently during model adaptation \cite{ding2023parameter,wengbitfit+}. However, the \baichuanchange{link} between \baichuanchange{fine-tuning different bias terms} and downstream performance during model adaptation is not clear \cite{ding2023parameter,wengbitfit+}. \baichuanchange{Moreover, the selective \gls{PEFT} approaches, such as the magnitude of bias change \cite{ansell2022composable} and empirical Fisher information \cite{sung2021training,xue2025fish,guozeroth}, provide limited guidance to select the target bias term.}
As a result, strategies for selecting target bias terms for effective fine-tuning remain largely unexplored.

\section{Conclusions} 
Bias-only fine-tuning \baichuanchange{of \glspl{LLM}} provides out-of-the-box usability, and demonstrates competitive performance compared to full-parameter fine-tuning, particularly in low-data regimes \cite{zaken2022bitfit,logan2022cutting,doering2024empirical}. In this paper, we introduce a simple yet effective approach, \aclbaichuan{based on the potential expressive power of bias terms in query, key, and value projections,}
to investigate the \baichuanchange{target} bias term for effective fine-tuning. We extensively evaluate our approach and shed light on the link between fine-tuning different bias terms and downstream performance, across diverse data regimes. Our key finding is that directly fine-tuning $\boldeq{b}_v$ generally leads to higher downstream performance in low-data regimes, compared to $\boldeq{b}_q$ and $\boldeq{b}_k$, without requiring any post-hoc evaluation. Moreover, we extend and generalize our key finding to various \glspl{LLM} covering encoder-only (masked) and decoder-only (autoregressive) \glspl{LLM} \cite{vaswani2017attention} with sizes ranging from 110M to 6.7B parameters, as well as bias-free \glspl{LLM}, across various downstream tasks, including classification, multiple-choice, and generation tasks. At the same time, our work is compatible and can be readily combined with \gls{PEFT} methods, e.g., \gls{LoRA}, \gls{VeRA}, \gls{DoRA}, \camera{and PiSSA}. Our results show the effectiveness of directly fine-tuning $\boldeq{b}_v$, without any post-hoc evaluation.

\section*{Limitations}
\addcontentsline{toc}{section}{Limitations}
In this work, we shed light on the importance of fine-tuning bias terms in \glspl{LLM} for unprecedented parameter efficiency. The limitations could be summarized as follows: (1) Our paper mainly focuses on standard softmax attention, not including linear attention \cite{katharopoulos2020transformers,choromanski2021rethinking}; (2) Our paper mainly focuses on the query, key, and value projections for \gls{SDPA}, similar to previous work \cite{qiu2025gated}. 
\aclbaichuan{Recent studies have shown that LayerNorm can potentially be replaced by a simple dynamic Tanh function \cite{zhu2025transformers} and even further by a rescaled Gaussian cumulative distribution function \cite{chen2025strongernormalizationfreetransformers}. At the same time, the feed-forward network can potentially be replaced by efficient non-linear mapping into low-dimensional computation \cite{kim2025eugens}; 
(3) Our paper analyzes the improvement in expressiveness brought by $\boldeq{b}_q$, $\boldeq{b}_k$, and $\boldeq{b}_v$} in Section \ref{sec:method}. How these bias terms influence the degree of superposition \cite{liu2025superposition} and, consequently, the scaling law \cite{kaplan2020scaling}, is left for future investigation.

\section*{Acknowledgments}
This research has been partially supported by the Swedish Wallenberg AI, Autonomous Systems and Software Program (WASP), Swedish Research Council, Swedish Foundation for Strategic Research, ELLIIT Strategic Research Environment, and an unrestricted gift from Google. 
\bibliography{custom}

\appendix

\clearpage

\section*{Appendix}
\addcontentsline{toc}{section}{Appendix}
\tableofcontents

\vspace{40pt}
\section{Different Approaches of Investigating Bias Terms}
\subsection{Magnitude Approach}
\label{sec:appendix_magnitude}
We denote \baichuanchange{the bias term at layer $l$} before fine-tuning as $\boldeq{b}_{\mathcal{T}}^{(l),pre}$ and after fine-tuning as $\boldeq{b}_{\mathcal{T}}^{(l),post}$. The change of the bias term across all layers in Magnitude approaches \cite{zaken2022bitfit} is denoted as:
\begin{align}
\Delta(\boldeq{b}_{\mathcal{T}}) = \frac{1}{L}\sum_{l=1}^{L} \left \| \boldeq{b}_{\mathcal{T}}^{(l),post} - \boldeq{b}_{\mathcal{T}}^{(l),pre} \right \|_1. \nonumber
\end{align}

\subsection{Fisher Approach}
\label{sec:appendix_fisher}

For the Fisher information, \baichuanchange{the effect of changes in the bias before fine-tuning ($\boldeq{b}_{\mathcal{T}}^{pre}$) on the output $\boldeq{y}$ is measured, given training data $\boldeq{x}$}. Fisher information is typically approximated using its diagonal for computational efficiency, giving rise to the empirical Fisher information applied in supervised learning.

\begin{figure*}[!ht]
  \centering
  \begin{subfigure}{0.47\textwidth}
    \includegraphics[width=\linewidth]{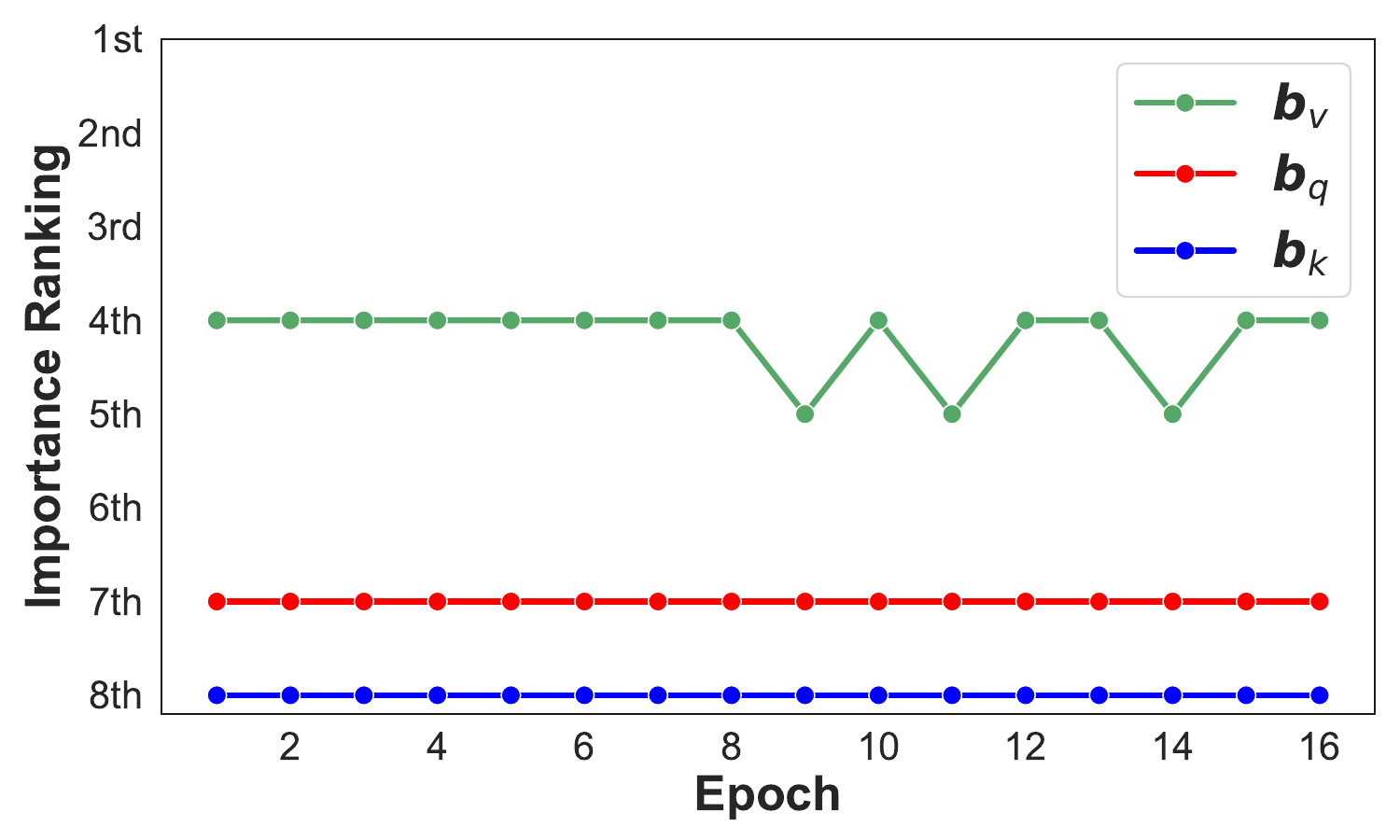}
    \caption{Batch size 16}
  \end{subfigure}
  \begin{subfigure}{0.47\textwidth}
    \includegraphics[width=\linewidth]{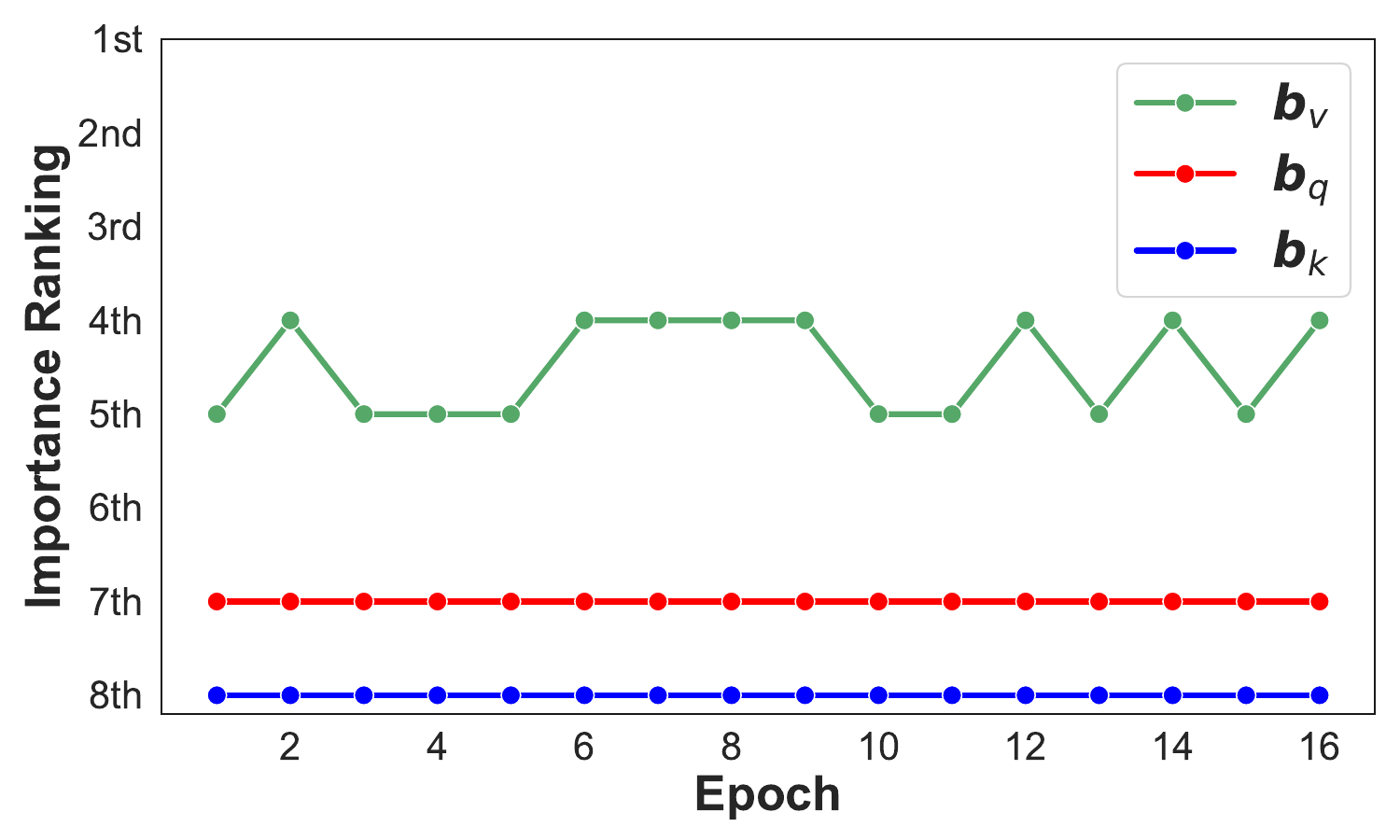}
    \caption{Batch size 8}
  \end{subfigure}
  \caption{Fisher Approach: importance ranking of different bias terms when fine-tuning BERT$_{\mathrm{BASE}}$ on the SST2 low-data regime as epoch increasing. Although different batch sizes introduce slight variations to the importance ranking of $\boldeq{b}_v$, the underlying pattern (mostly static) remains consistent.}
  \label{fig:appendix_fisher_epoch}
  \vspace{20pt}
\end{figure*}

\begin{figure*}[!ht]
\centering
  \begin{subfigure}[t]{0.32\textwidth}
    \includegraphics[width=\linewidth]{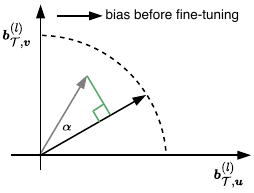}
    \caption{$\boldeq{b}^{(l),post}_{\mathcal{T}}<\boldeq{b}^{(l),pre}_{\mathcal{T}}$}
  \end{subfigure}
  \begin{subfigure}[t]{0.32\textwidth}
    \includegraphics[width=\linewidth]{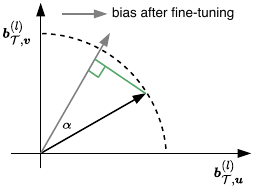}
\caption{$\boldeq{b}^{(l),post}_{\mathcal{T}}>\boldeq{b}^{(l),pre}_{\mathcal{T}}$}
  \end{subfigure}
  \begin{subfigure}[t]{0.32\textwidth}
  \raisebox{2.2pt}{
    \includegraphics[width=\linewidth]{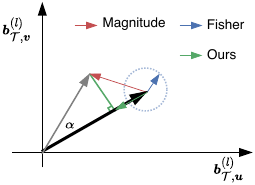}
    }\caption{Comparison}
  \end{subfigure}
  \caption{We propose a simple yet effective approach, incorporating angular changes $\alpha$ into magnitude changes, to measure the change of \baichuanchange{$\boldeq{b}_{\mathcal{T}}^{(l)}$} by projection and scaling (simplified visualization in $(\boldeq{u},\boldeq{v})\in\mathbb{R}^2$).}
  \label{fig:method}
\end{figure*}

For a \gls{LLM} model with pre-trained parameters $\boldeq{\theta} \in \mathbb{\mathbb{R}}^{|\boldeq{\theta}|}$, the output $\boldeq{y}$ is inferred by the input $\boldeq{x}$ and the parameters $\boldeq{\theta}$, i.e., $\boldeq{y} = f_{\boldeq{\theta}}(\boldeq{x})$. To measure how much the prediction would change for a given change in parameters $\boldeq{\theta}$, the Fisher information $F_{\boldeq{\theta}}\in \mathbb{R}^{|\boldeq{\theta}|\times|\boldeq{\theta}|}$ is introduced as below,
\resizebox{\columnwidth}{!}{$
\begin{aligned}
F_{\boldeq{\theta}} =
\mathbb{E}_{\boldeq{x}\sim p(\boldeq{x})}\left [ \mathbb{E}_{\boldeq{y}\sim p_{\boldeq{\theta}}(\boldeq{y}\mid \boldeq{x})}\nabla_{\boldeq{\theta}}\log p_{\boldeq{\theta}}(\boldeq{y}|\boldeq{x}){\nabla_{\boldeq{\theta}}\log p_{\boldeq{\theta}}(\boldeq{y}|\boldeq{x})}^T \right ], \nonumber
\end{aligned}$}
where $p(\boldeq{x})$ is the probability distribution of the input $\boldeq{x}$ and $p_{\boldeq{\theta}}(\boldeq{y}|\boldeq{x})$ is the probability distribution of the output $\boldeq{y}$ based on the \gls{LLM} model parameters $\boldeq{\theta}$ and input $\boldeq{x}$. $\nabla \log p_{\boldeq{\theta}}(\boldeq{y}|\boldeq{x})$ is the gradient of the log-likelihood with respect to $\theta$. In \glspl{LLM}, $|\boldeq{\theta}|\times|\boldeq{\theta}|$ has heavy computation, making it impossible to compute. Therefore, the diagonal vector is exploited as the approximated Fisher information. Furthermore, due to the limited training data, the approximated Fisher information $\hat{F}_{\boldeq{\theta}} \in \mathbb{R}^{|\boldeq{\theta}|}$ is introduced as below:
\begin{align}
\hat{F}(\boldeq{\theta} ) = \frac{1}{N}\sum_{i=1}^{N}  \mathbb{E}_{\boldeq{y}\sim p_{\boldeq{\theta}}(\boldeq{y}\mid \boldeq{x}_i)}(\nabla_{\boldeq{\theta}}\log p_{\boldeq{\theta}}(\boldeq{y}|\boldeq{x}_i))^2 , \nonumber
\end{align}
where $N$ is the total number of training data. Moreover, in supervised learning, we can use empirical Fisher information as below:
\begin{align}
\hat{F}(\boldeq{\theta} ) = \frac{1}{N}\sum_{i=1}^{N}  (\nabla_{\boldeq{\theta}}\log p_{\boldeq{\theta}}(\boldeq{y}_i|\boldeq{x}_i))^2.\nonumber
\end{align}
In supervised learning, this gradient $\nabla_{\theta}\log p_{\boldeq{\theta}}(\boldeq{y}_i|\boldeq{x}_i) $ is simplified as $-\nabla_{\theta}\mathcal{L}(\boldeq{y}_i,f_{\boldeq{\theta}}(\boldeq{x}_i)) $, where $\mathcal{L}$ is the Cross-Entropy loss.

Therefore, the change of the bias term across all layers in the Fisher approach is denoted as follows:
\begin{align}
\Delta(\boldeq{b}_{\mathcal{T}}) =  \frac{1}{L\cdot N}\sum_{l=1}^{L}\sum_{i=1}^{N}\left(\nabla_{\boldeq{b}^{(l),pre}_{\mathcal{T}}} \log p_{\theta}(\boldeq{y}_i|\boldeq{x}_i) \right)^2 , \nonumber
\end{align}
where $N$ is the total number of training data. $p_{\boldeq{\theta}}(\boldeq{y}_i|\boldeq{x}_i)$ is the probability distribution of the output $\boldeq{y}_i$ based on the \gls{LLM} model parameters $\boldeq{\theta}$ and input $\boldeq{x}_i$. $\nabla_{\boldeq{b}^{(l),pre}_{\mathcal{T}}}\log p_{\boldeq{\theta}}(\boldeq{y}_i|\boldeq{x}_i) $ is the gradient of the log-likelihood with respect to the bias $\boldeq{b}^{(l),pre}$.

\subsection{Our Approach}
\label{sec:appendix_projection_ratio}

When the magnitude of fine-tuned bias is smaller than the magnitude of the bias before fine-tuning, as illustrated in Fig. \ref{fig:method} (a), the projection of fine-tuned bias onto the bias before fine-tuning is considered to calculate the $\Delta(\boldeq{b}_{\mathcal{T}}^{(l)})$:
\resizebox{\columnwidth}{!}{$
\begin{aligned}
\Delta(\boldeq{b}_{\mathcal{T}}^{(l)}) =1- \frac{\| \boldeq{b}^{(l),post}_{\mathcal{T}} \|_2 \cdot \cos(\alpha)}{\| \boldeq{b}^{(l),pre}_{\mathcal{T}} \|_2} 
= 1- \frac{\boldeq{b}^{(l),pre}_{\mathcal{T}}\cdot \boldeq{b}^{(l),post}_{\mathcal{T}}}{\| \boldeq{b}^{(l),pre}_{\mathcal{T}} \|^2_2},   \nonumber
\end{aligned}$}
where $\alpha$ is the angle between the bias before and after fine-tuning. On the contrary, when the magnitude of fine-tuned bias is not smaller than the magnitude of the bias before fine-tuning, as illustrated in Fig. \ref{fig:method} (b), the projection of the bias before fine-tuning onto the fine-tuned bias is considered to calculate the $\Delta(\boldeq{b}_{\mathcal{T}}^{(l)})$:
\resizebox{\columnwidth}{!}{$
\begin{aligned}
\Delta(\boldeq{b}_{\mathcal{T}}^{(l)}) =1- \frac{\| \boldeq{b}^{(l),pre}_{\mathcal{T}} \|_2 \cdot \cos(\alpha)}{\| \boldeq{b}^{(l),post}_{\mathcal{T}} \|_2} 
= 1- \frac{\boldeq{b}^{(l),pre}_{\mathcal{T}}\cdot \boldeq{b}^{(l),post}_{\mathcal{T}}}{\| \boldeq{b}^{(l),post}_{\mathcal{T}} \|^2_2}.   \nonumber
\end{aligned}$}
To present a unified formulation considering all layers, we have:
\resizebox{\columnwidth}{!}{$
\begin{aligned}
\Delta(\boldeq{b}_{\mathcal{T}})= 
\frac{1}{L}\sum_{l=1}^{L}\left (1- \frac{\boldeq{b}^{(l),pre}_{\mathcal{T}}\cdot \boldeq{b}^{(l),post}_{\mathcal{T}}}{\max \left( \| \boldeq{b}^{(l),pre}_{\mathcal{T}} \|^2_2,\| \boldeq{b}^{(l),post}_{\mathcal{T}}  \|^2_2\right)} \right ). \nonumber
\end{aligned}$}

\newpage
\section{Extension of Analysis}

\subsection{Inherent Sharpness and High Sparsity of Attention Weights}
\label{sec:appendix_sharpe_sparse}

\begin{figure}[!ht]
  \centering
  \begin{subfigure}{0.33\textwidth}
    \includegraphics[width=\linewidth]{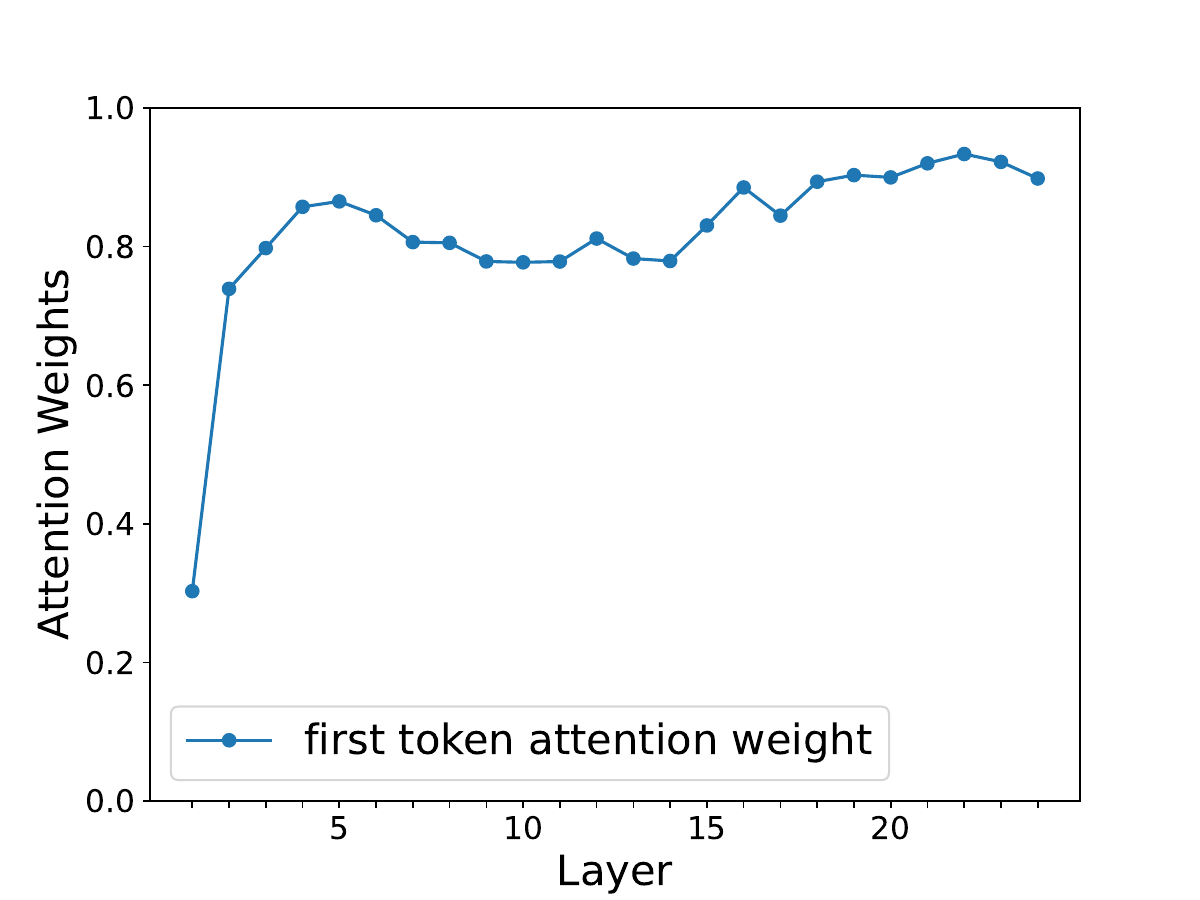}
    \caption{Proportion of attention allocated to the first token per layer (OPT-1.3B).}
  \end{subfigure}
  \begin{subfigure}{0.3\textwidth}
    \includegraphics[width=\linewidth]{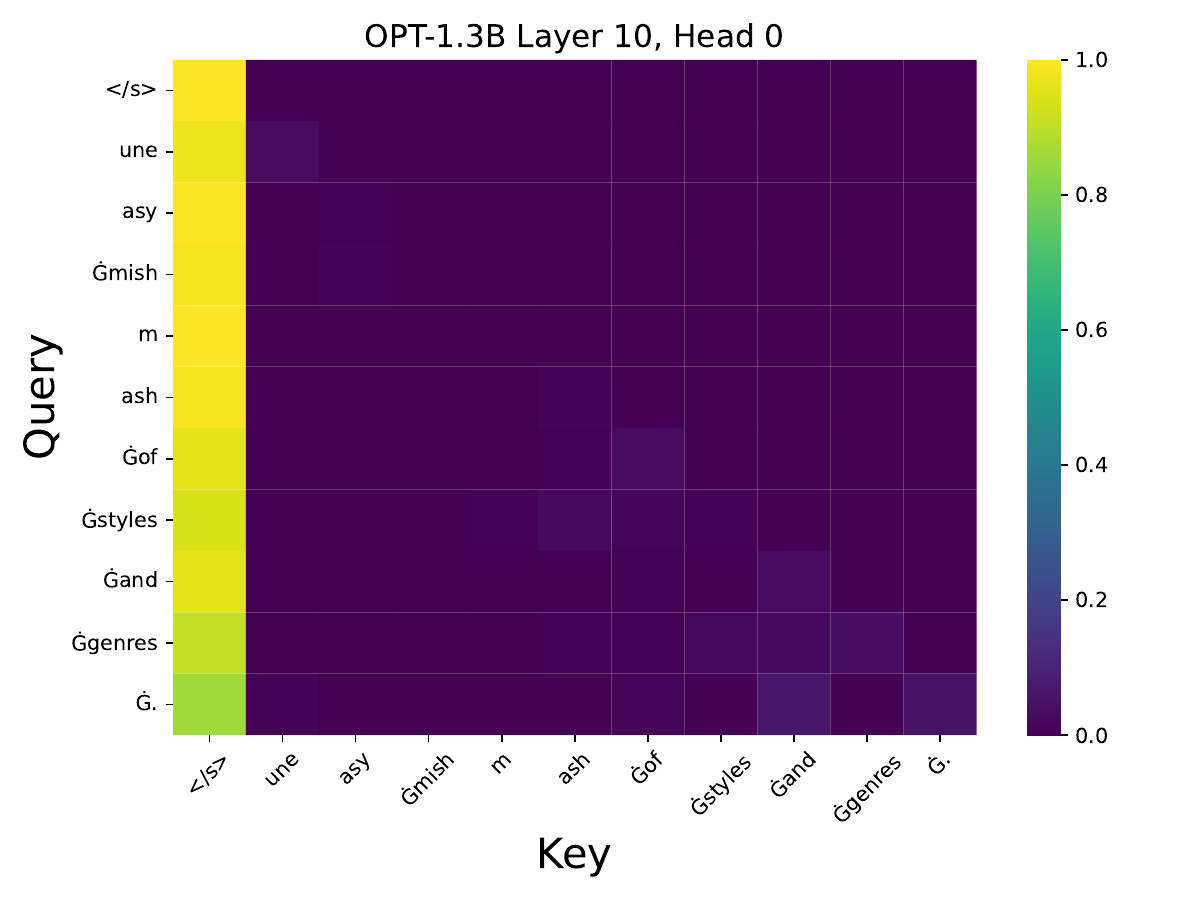}
    \caption{Attention weights for one head of the 10th layer (OPT-1.3B).}
  \end{subfigure}
  \begin{subfigure}{0.3\textwidth}
    \includegraphics[width=\linewidth]{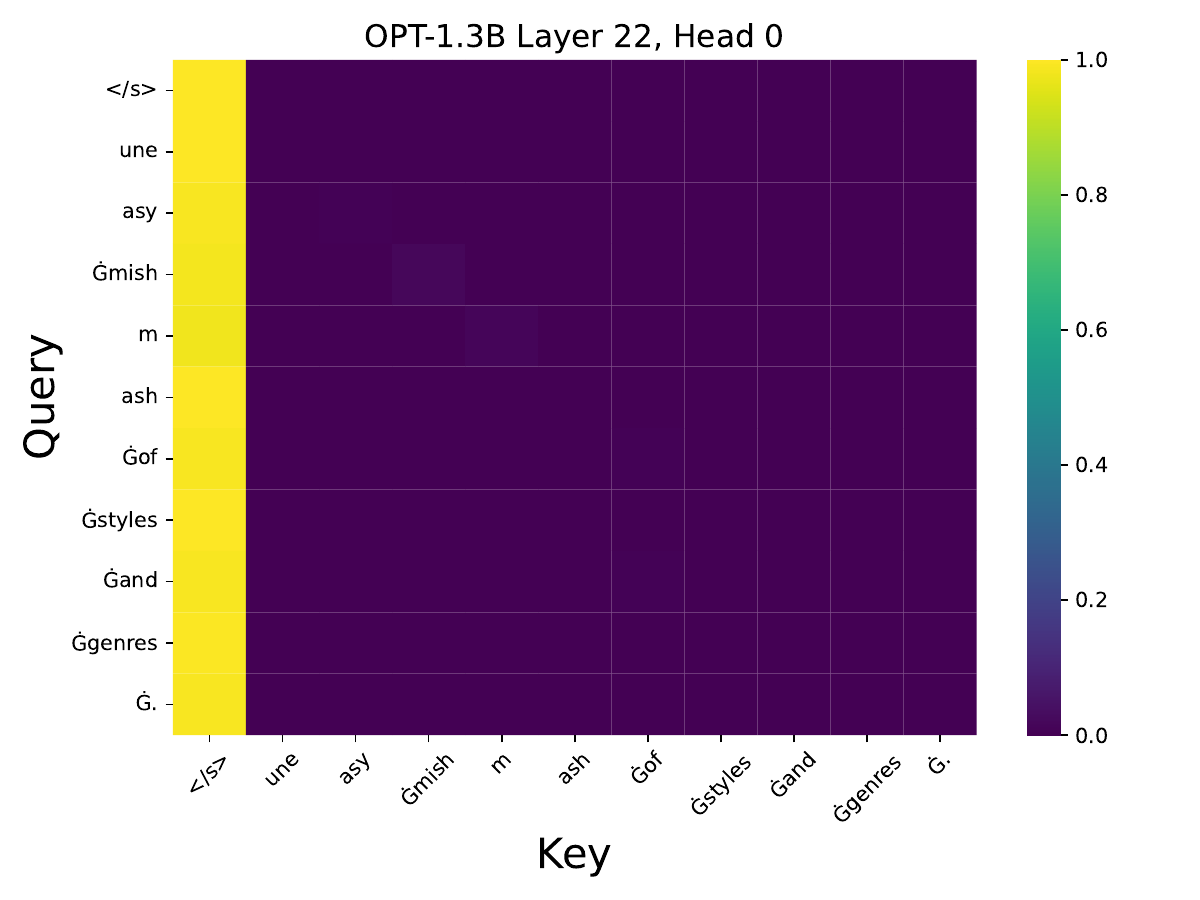}
    \caption{Attention weights for one head of the 22nd layer (OPT-1.3B).}
  \end{subfigure}
  \begin{subfigure}{0.3\textwidth}
    \includegraphics[width=\linewidth]{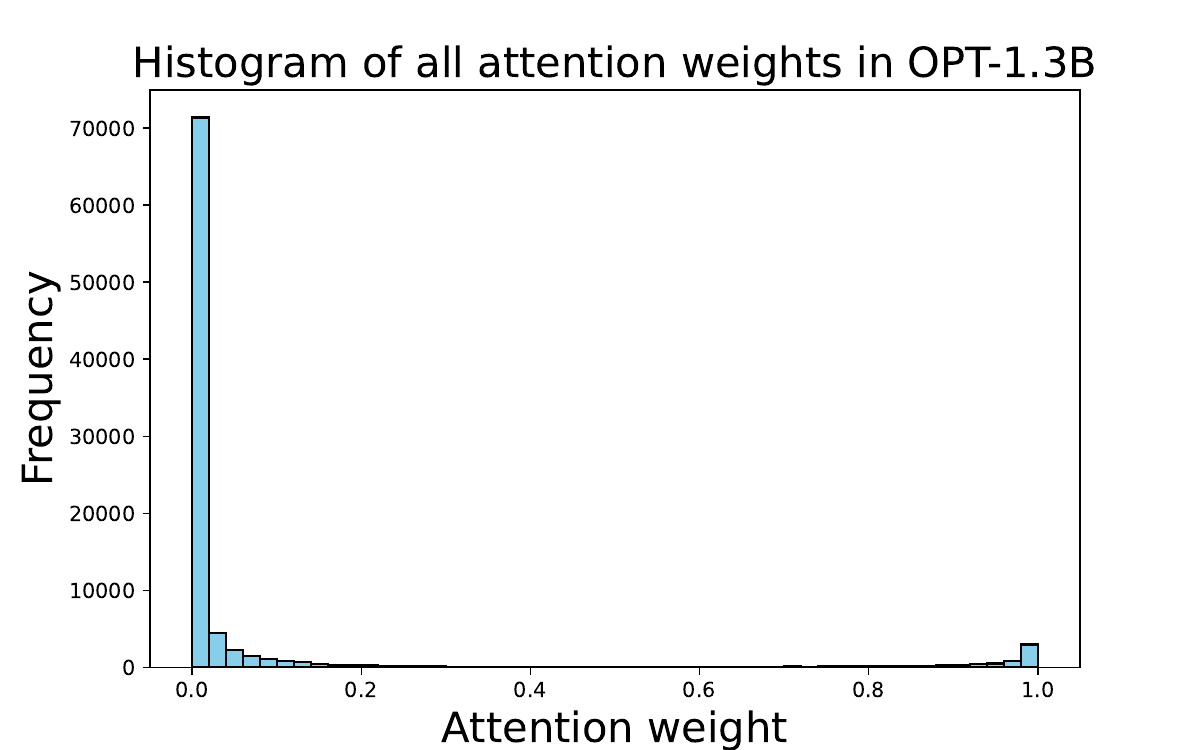}
    \caption{The histograms of all attention weights (OPT-1.3B).}
  \end{subfigure}
  \caption{The observed attention sink \cite{xiao2024efficient,qiu2025gated} and the histograms of all attention weights for OPT-1.3B (unidirectional attention) with one test input from the SST-2—``\textit{uneasy mishmash of styles and genres .}''. In OPT-1.3B, attention weights across layers tend to be directed towards the first token and show high sparsity.
} 
  \label{fig:appendix_attention_sink}
\end{figure}

\newpage
\vspace*{51pt}
\begin{figure}[H]
  \centering
  \begin{subfigure}{0.3\textwidth}
    \includegraphics[width=\linewidth]{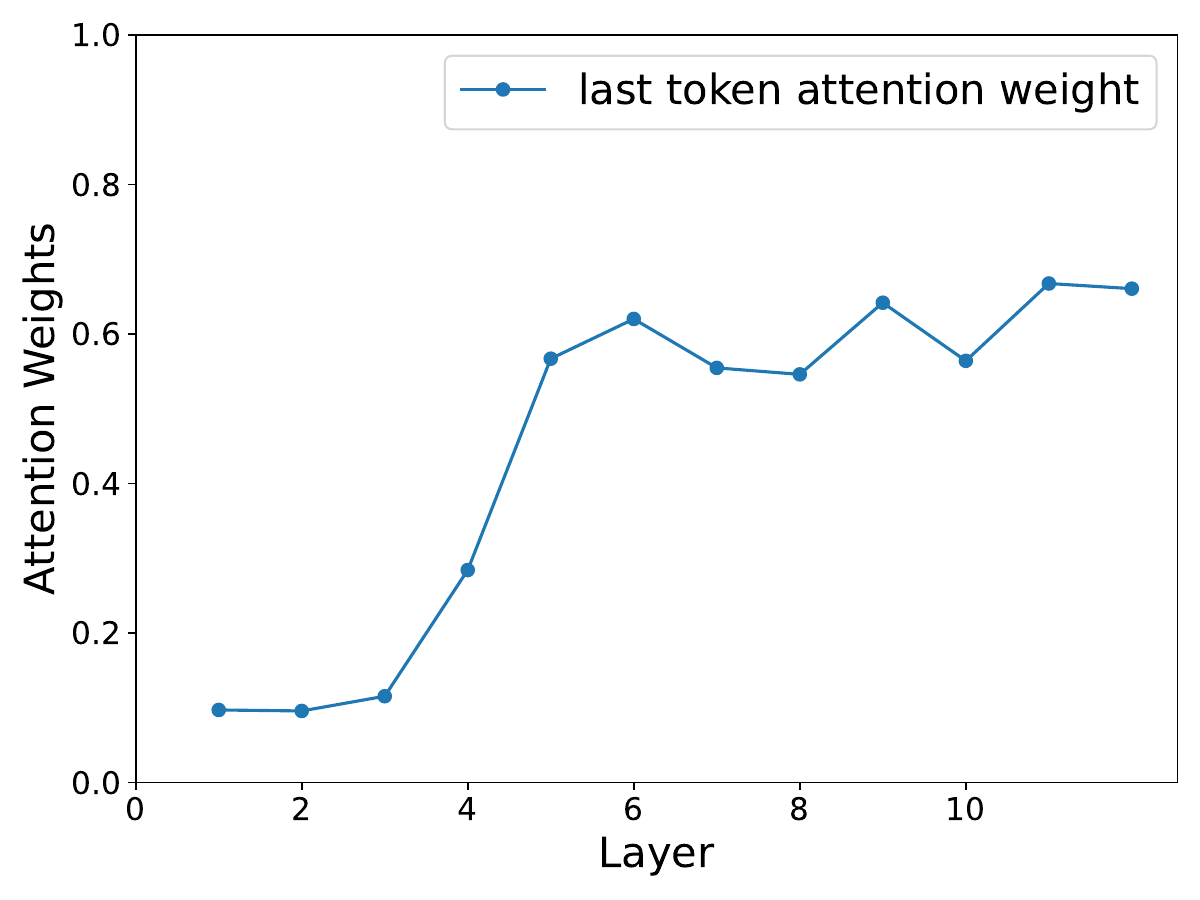}
    \caption{Proportion of attention allocated to the last token per layer (BERT$_{\mathrm{BASE}}$).}
  \end{subfigure}
  \begin{subfigure}{0.3\textwidth}
    \includegraphics[width=\linewidth]{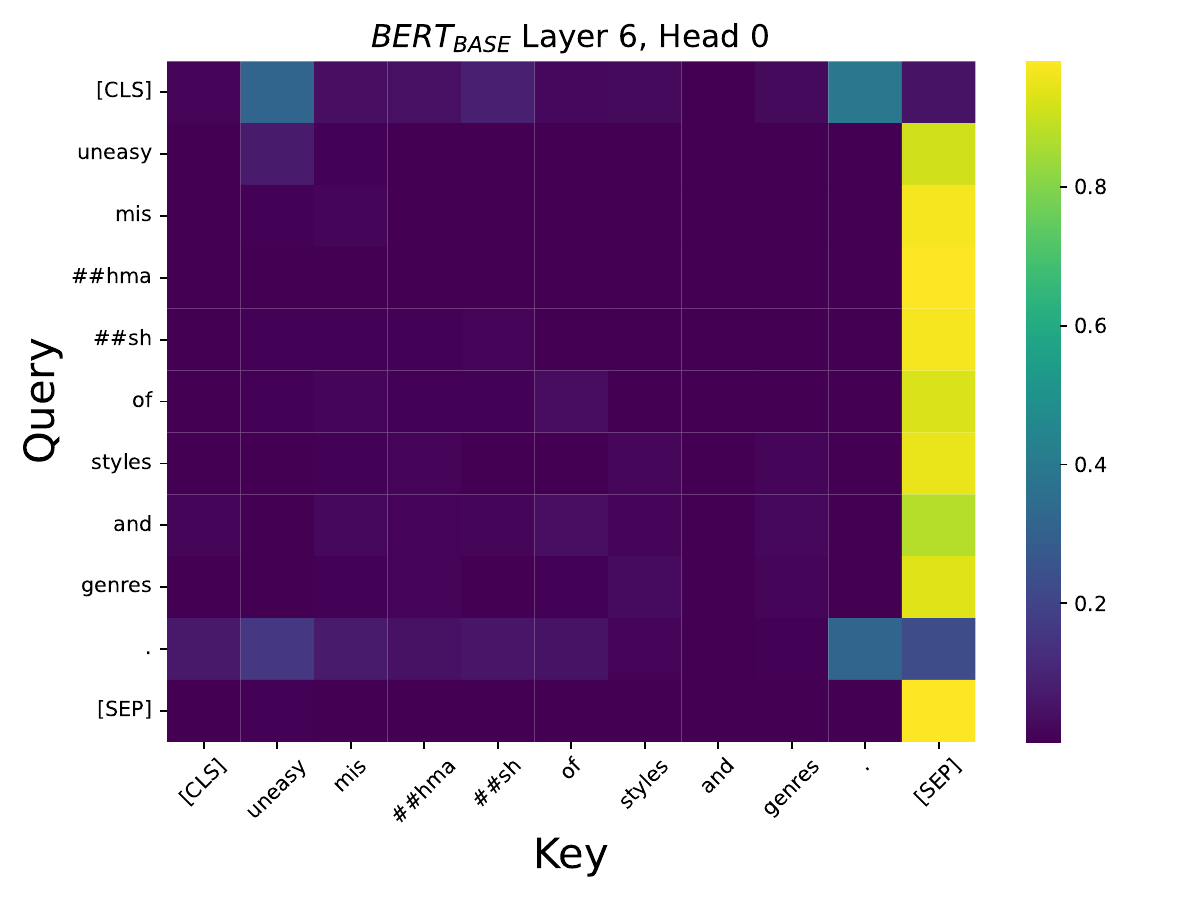}
    \caption{Attention weights for one head of the 6th layer (BERT$_{\mathrm{BASE}}$).}
  \end{subfigure}
  \begin{subfigure}{0.3\textwidth}
    \includegraphics[width=\linewidth]{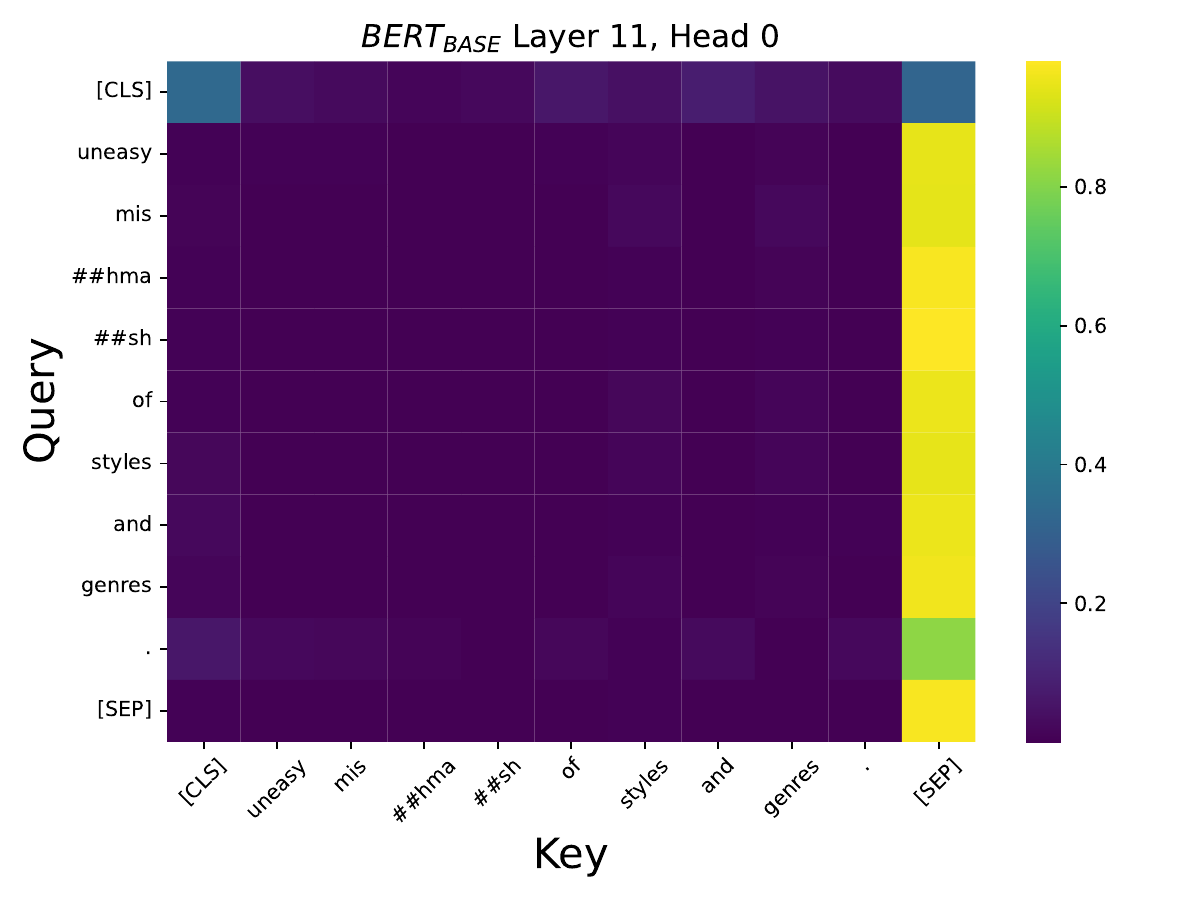}
    \caption{Attention weights for one head of the 11st layer (BERT$_{\mathrm{BASE}}$).}
  \end{subfigure}
  \begin{subfigure}{0.3\textwidth}
    \includegraphics[width=\linewidth]{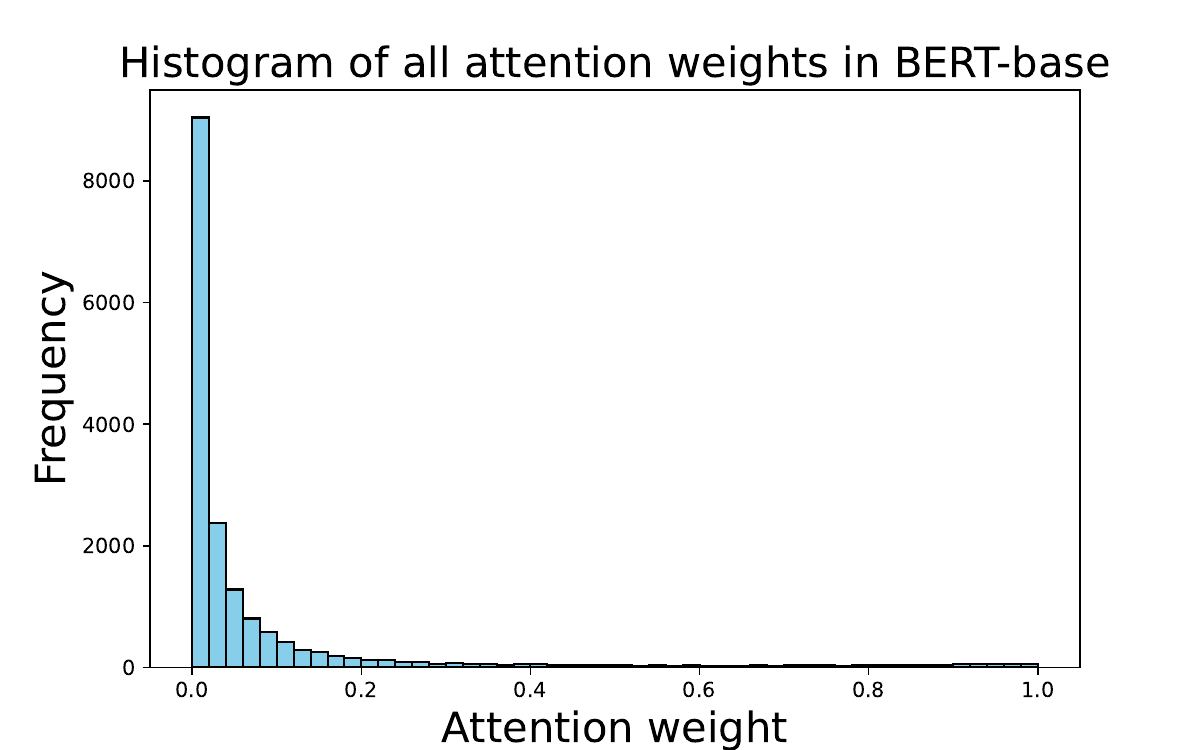}
    \caption{The histograms of all attention weights (BERT$_{\mathrm{BASE}}$).}
  \end{subfigure}
  \caption{Similar attention sink and the histograms of all attention weights for BERT$_{\mathrm{BASE}}$ (bidirectional attention) with one test input from the SST-2—``\textit{uneasy mishmash of styles and genres .}''. In BERT$_{\mathrm{BASE}}$, the attention sink happens for the last token and show high sparsity.}
  \label{fig:appendix_histogram}
\end{figure}

\onecolumn
\subsection{Equivalence of Bias Terms in the Value Projection and Output Projection}
\label{sec:appendix_sdpa_with_output}

Based on Equation (\ref{eq:bv}), if only $\boldeq{b}_v$ available, for the whole attention weights $\boldeq{A}$ of each attention head, we have:
\begin{align}
\mathrm{head}^i=\boldeq{A}^i(\boldeq{V}^i+\boldeq{b}_v^i)=\boldeq{A}^i\boldeq{V}^i+\boldeq{b}_v^i, \label{eq:b_v_A}
\end{align}
where $i$ is the index of attention head; $\boldeq{A}^i\boldeq{V}^i \in \mathbb{R}^{n\times d_k}$ and $\boldeq{b}_v^i \in \mathbb{R}^{1\times d_k}$. $\boldeq{b}_v^i $ is broadcast across the rows of $\boldeq{A}^i\boldeq{V}^i $. For multi-head attention, Equation (\ref{eq:b_v_A}) is repeated in parallel for $h$ heads and $h=\frac{d_{\mathrm{model}}}{d_k}$, where $d_{\mathrm{model}}$ is the model dimension and $d_k$ is the dimension of one head.

The outputs of \gls{SDPA} are concatenated as:
\begin{align}
\mathrm{MultiHead}=\mathrm{Concat}(\mathrm{head}^1,...,\mathrm{head}^h), \nonumber
\end{align}
where the $\mathrm{MultiHead} \in \mathbb{R}^{n\times d_{\mathrm{model}}}$ is passed through the output projection as shown in Fig. \ref{fig:appendix_SDPA_with_output}. For the output projection, we have $\boldeq{W}_o \in \mathbb{R}^{d_{\mathrm{model}}\times d_{\mathrm{model}}}$ and $\boldeq{b}_o \in \mathbb{R}^{1\times d_{\mathrm{model}}}$. If only $\boldeq{b}_v$ available, the calculation is:
\begin{align}
\mathrm{MultiHead}_{(\mathrm{with}\;\boldeq{b}_v)} \boldeq{W}_o&=\mathrm{Concat}(\mathrm{head}^1,...,\mathrm{head}^h)\boldeq{W}_o  \label{eq:bv_1} \\
& = \mathrm{Concat}(\boldeq{A}^1\boldeq{V}^1+\boldeq{b}_v^1,...,\boldeq{A}^h\boldeq{V}^h+\boldeq{b}_v^h)\boldeq{W}_o \nonumber \\
& = \left [\mathrm{Concat}(\boldeq{A}^1\boldeq{V}^1,...,\boldeq{A}^h\boldeq{V}^h) + \mathrm{Concat}(\boldeq{b}_v^1,...,\boldeq{b}_v^h) \right ] \boldeq{W}_o \nonumber \\
& = \mathrm{Concat}(\boldeq{A}^1\boldeq{V}^1,...,\boldeq{A}^h\boldeq{V}^h)\boldeq{W}_o + \mathrm{Concat}(\boldeq{b}_v^1,...,\boldeq{b}_v^h)\boldeq{W}_o, \nonumber
\end{align}
where $\mathrm{Concat}(\boldeq{A}^1\boldeq{V}^1,...,\boldeq{A}^h\boldeq{V}^h) \in \mathbb{R}^{n\times d_{\mathrm{model}}}$ and $\mathrm{Concat}(\boldeq{b}_v^1,...,\boldeq{b}_v^h) \in \mathbb{R}^{1\times d_{\mathrm{model}}}$. $\mathrm{Concat}(\boldeq{b}_v^1,...,\boldeq{b}_v^h) $ is broadcast across the rows of $\mathrm{Concat}(\boldeq{A}^1\boldeq{V}^1,...,\boldeq{A}^h\boldeq{V}^h)$.

If only $\boldeq{b}_o$ available, the calculation is: 
\begin{align}
\mathrm{MultiHead}_{(\mathrm{without}\;\boldeq{b}_v)}\boldeq{W}_o + \boldeq{b}_o&=\mathrm{Concat}(\mathrm{head}^1,...,\mathrm{head}^h)\boldeq{W}_o  + \boldeq{b}_o \label{eq:bv_2} \\
& = \mathrm{Concat}(\boldeq{A}^1\boldeq{V}^1,...,\boldeq{A}^h\boldeq{V}^h)\boldeq{W}_o + \boldeq{b}_o. \nonumber \\
\nonumber
\end{align}

Based on Equation (\ref{eq:bv_1}) and Equation (\ref{eq:bv_2}), let $\boldeq{b}_{o,j}$ denote the $j$-th elements of $\boldeq{b}_o$ and $\boldeq{W}_{o,j}$ denote the $j$-th column of $\boldeq{W}_o$. When
\begin{align}
\boldeq{b}_{o,j} = \mathrm{Concat}(\boldeq{b}_v^1,...,\boldeq{b}_v^h)\boldeq{W}_{o,j}
\quad \forall j \in \left \{  1,...,d_{\mathrm{model}}\right \}, \nonumber
\end{align}
we can get
\begin{align}
\mathrm{MultiHead}_{(\mathrm{with}\;\boldeq{b}_v)} \boldeq{W}_o = \mathrm{MultiHead}_{(\mathrm{without}\;\boldeq{b}_v)}\boldeq{W}_o + \boldeq{b}_o. \nonumber
\end{align}
This derivation demonstrates the equivalence of bias terms in the value projection and output projection. In terms of improving the expressiveness, $\boldeq{b}_v$ is able to yield an effect equivalent to that of $\boldeq{b}_o$. We also empirically fine-tuning $\boldeq{b}_v$ and $\boldeq{b}_o$, and validate the claimed equivalence, as presented in Table \ref{tab:appendix_b_o_vs_b_v}.

\begin{minipage}{0.48\textwidth}
    \centering
    \includegraphics[width=0.4\linewidth]{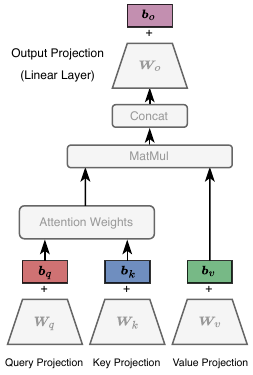}
    \captionof{figure}{The SDPA with output projection, where the output projection is a linear layer.}\label{fig:appendix_SDPA_with_output}
\end{minipage}
\hfill
\begin{minipage}{0.48\textwidth}
    \centering
    \vspace{1.8em}
    \begin{tabular}{ccc}
    \toprule[2pt]
     Multi-Seed&  $\boldeq{b}_v$ &  $\boldeq{b}_o$ \\
    \midrule[1pt]
    0 & 93.1 & 93.1 \\ 
    1 & 93.6 & 92.8 \\ 
    2 & 93.6 & 93.7 \\ 
    3 &92.4&92.3\\
    4 &93.2&93.1\\
    \hline
    mean ± std & 93.1\text{\scriptsize$\pm$0.44} & 93.0\text{\scriptsize$\pm$0.45}\\
    \bottomrule[2pt]
    \end{tabular}
    \captionof{table}{Downstream performance (\%) of fine-tuning $\boldeq{b}_v$ and $\boldeq{b}_o$ on SST-2 dataset with OPT-1.3B.}\label{tab:appendix_b_o_vs_b_v}
\end{minipage}

\twocolumn

\pagebreak

\section{Investigation of Extensions}
\label{sec:appendix_investigation}

\subsection{The Effectiveness of Our Key Finding}
\label{sec:appendix_qualitative_observations}

\begin{figure}[!ht]
  \centering
  \begin{subfigure}{0.45\textwidth}
    \includegraphics[width=\linewidth]{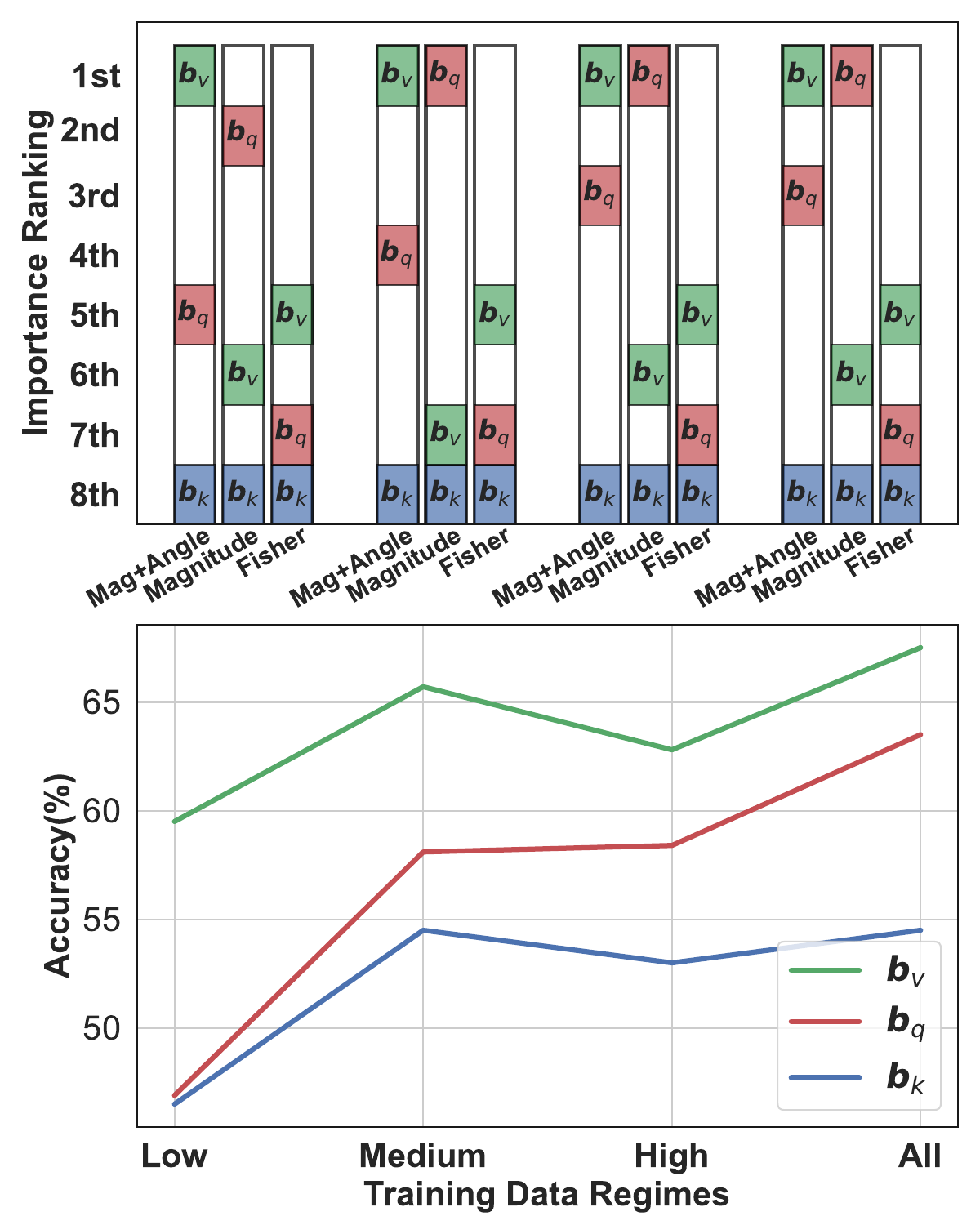}
    \caption{RTE}
  \end{subfigure}
  \caption{Importance ranking and downstream performance of fine-tuning different bias terms on the RTE dataset with BERT$_{\mathrm{BASE}}$. Still, our approach demonstrates the ability to precisely and dynamically identify the \baichuanchange{target} bias term across low-data to high-data regimes, outperforming both Magnitude and Fisher approaches.}
  \label{fig:appendix_rank_comparison}
\end{figure}

\begin{figure}[!ht]
  \centering
    \includegraphics[width=0.45\textwidth]{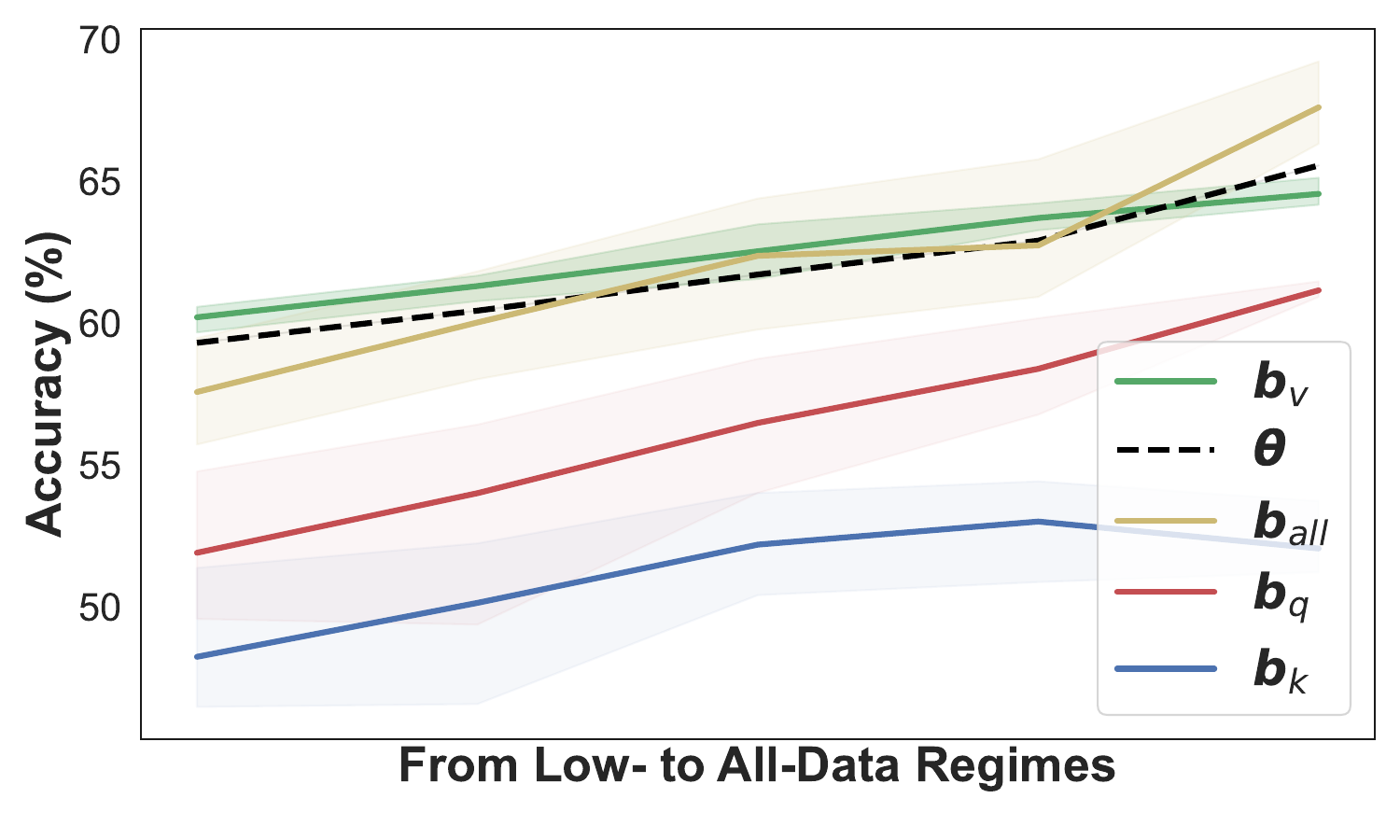}
    \caption{Fine-tuning various subsets of \gls{LLM} parameters. The solid line reports the mean over three different seeds, and the shaded area indicates the standard deviation. \baichuanchange{We report the last epoch accuracy on validation dataset for consistent comparison, while the BitFit \cite{zaken2022bitfit} reports the best validation accuracy across training for fine-tuning $\boldeq{b}_{all}$.}}
  \label{fig:comparison_rte}
\end{figure}

\begin{figure}[!ht]
\vspace{20pt}
  \centering
 \includegraphics[width=0.47\textwidth]{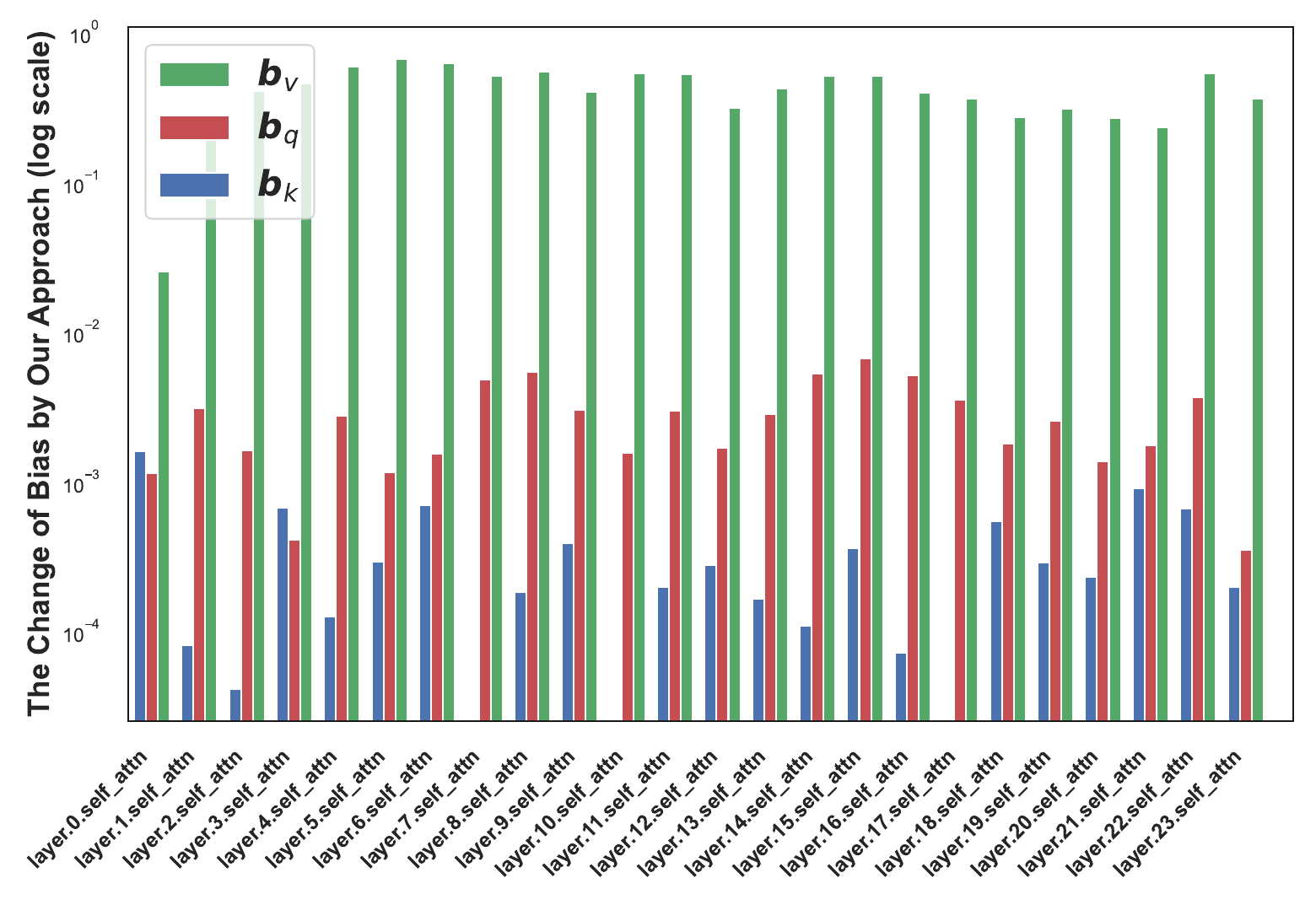}
    \caption{The change of bias (log scale), across Transformer layers, by our approach with OPT-1.3B on SST-2. The target bias is $\boldeq{b}_v$.}
  \label{fig:opt_target_bias}
\end{figure}

\subsection{Statistical Significance}
\label{sec:appendix_significance}
We did the paired t-test for the main results in Table \ref{tab:main} and Table \ref{tab:opt}. For Table \ref{tab:main}, we pick CoLA and STS-B to do paired t-test, where the results are statistically significant ($p$<0.05), as shown below.
\begin{table}[!ht]
\centering
\begin{tabular}{ccc}
\toprule[2pt]
 & CoLA & STS-B \\
\midrule[1pt]
Fine-Tuning $\boldeq{b}_v$ &  40.2\text{\scriptsize$\pm$2.85}&75.3\text{\scriptsize$\pm$0.65}  \\ 
t-test ($\boldeq{b}_v$ vs $\boldeq{b}_q$) & $p$=0.005 & $p$= 0.001 \\
Fine-Tuning $\boldeq{b}_q$ &  27.5\text{\scriptsize$\pm$3.98}&67.5\text{\scriptsize$\pm$0.28}  \\ 
t-test ($\boldeq{b}_q$ vs $\boldeq{b}_k$) & $p$=0.003 & $p$=0.026  \\
Fine-Tuning $\boldeq{b}_k$ &  4.1\text{\scriptsize$\pm$2.93}&65.6\text{\scriptsize$\pm$0.48}  \\ 
\bottomrule[2pt]
\end{tabular}
\caption{Paired t-test shows the statistical significance of our results in Table \ref{tab:main}.}
\label{tab:appendix_p_test_1}
\end{table}
To further investigate this paired t-test, taking SST-2 for example, we repeat the experiments 5 times with different random seeds to obtain the $p$-values of $0.003$ and $0.013$, showing the statistical significance of these results. When we repeat the experiments 10 times with different random seeds, we obtain the $p$-values of $1.44\times10^{-6}$ and $2.78\times10^{-5}$, showing the high statistical significance.

For Table \ref{tab:opt}, we pick OPT-1.3B on CB, SQuAD, and DROP, where the results are statistically significant ($p$<0.05), as shown below:
\begin{table}[!ht]
\centering
\resizebox{\columnwidth}{!}{
\begin{tabular}{cccc}
\toprule[2pt]
 & CB & SQuAD & DROP\\
\midrule[1pt]
Fine-Tuning $\boldeq{b}_v$ &  77.3\text{\scriptsize$\pm$2.2}&78.6\text{\scriptsize$\pm$0.9}&28.8\text{\scriptsize$\pm$0.5}  \\ 
t-test ($\boldeq{b}_v$ vs $\boldeq{b}_q$) & $p$=0.023 & $p$= 0.0004 & $p$=0.021 \\
Fine-Tuning $\boldeq{b}_q$ &  70.2\text{\scriptsize$\pm$1.7}&68.5\text{\scriptsize$\pm$0.6}&22.4\text{\scriptsize$\pm$1.7}  \\ 
t-test ($\boldeq{b}_q$ vs $\boldeq{b}_k$) & $p$=0.001 & $p$=0.0001 & $p$=0.009  \\
Fine-Tuning $\boldeq{b}_k$ &  37.5\text{\scriptsize$\pm$0.0}&24.6\text{\scriptsize$\pm$0.3}&10.3\text{\scriptsize$\pm$0.1}  \\ 
\bottomrule[2pt]
\end{tabular}}
\caption{Paired t-test shows the statistical significance of our results in Table \ref{tab:opt}.}
\label{tab:appendix_p_test_2}
\end{table}

\color{black}

\begin{table*}[!ht]
\centering
\resizebox{\textwidth}{!}{
\begin{tabular}{ccccccccccccc}
\toprule[2pt]
 \multirow{2}{*}{LLMs}   & Fine-Tuning  & SST-2  & RTE & CB &BoolQ &WSC& WiC  & MultiRC & COPA & ReCoRD & SQuAD & DROP  \\
 &Techniques& \multicolumn{7}{c}{----------------------classification----------------------}& \multicolumn{2}{c}{--multiple choice--}&\multicolumn{2}{c}{---generation---}\\
\midrule[1pt]
\rowcolor{mygreen!40}
\cellcolor{white}
\multirow{3}{*}{OPT-1.3B} 
&  $\boldeq{b}_v$ &\textbf{93.4}\text{\scriptsize$\pm$0.2}&\textbf{70.5}\text{\scriptsize$\pm$1.1}&\textbf{77.3}\text{\scriptsize$\pm$2.2}&\textbf{66.4}\text{\scriptsize$\pm$0.9}&\textbf{63.4}\text{\scriptsize$\pm$0.1}&\textbf{61.8}\text{\scriptsize$\pm$1.2}&\textbf{63.8}\text{\scriptsize$\pm$3.8}&\textbf{77.6}\text{\scriptsize$\pm$0.5}&\textbf{71.2}\text{\scriptsize$\pm$0.7}&\textbf{78.6}\text{\scriptsize$\pm$0.9}&\textbf{28.8}\text{\scriptsize$\pm$0.5}  \\
 & $\boldeq{b}_q$ &73.5\text{\scriptsize$\pm$0.7}&56.4\text{\scriptsize$\pm$1.0}&70.2\text{\scriptsize$\pm$1.7}&60.8\text{\scriptsize$\pm$1.1}&62.8\text{\scriptsize$\pm$1.2}&58.9\text{\scriptsize$\pm$0.2}&56.6\text{\scriptsize$\pm$0.4}&74.0\text{\scriptsize$\pm$0.0}&70.6\text{\scriptsize$\pm$0.6}&68.5\text{\scriptsize$\pm$0.6}&22.4\text{\scriptsize$\pm$1.7} \\
 &  $\boldeq{b}_k$ &53.5\text{\scriptsize$\pm$0.0}&53.0\text{\scriptsize$\pm$0.1}&37.5\text{\scriptsize$\pm$0.0}&46.8\text{\scriptsize$\pm$1.6}&44.2\text{\scriptsize$\pm$0.0}&57.0\text{\scriptsize$\pm$0.0}&45.8\text{\scriptsize$\pm$1.2}&75.0\text{\scriptsize$\pm$0.0}&70.0\text{\scriptsize$\pm$0.3}&24.6\text{\scriptsize$\pm$0.3}&10.3\text{\scriptsize$\pm$0.1}  \\
 \hline
 %CB seed=1 lr=1e-3
 \rowcolor{mygreen!40}
\cellcolor{white}
 \multirow{3}{*}{OPT-6.7B} 
&  $\boldeq{b}_v$ &\textbf{95.4}\text{\scriptsize$\pm$0.3}&\textbf{81.4}\text{\scriptsize$\pm$0.8}&\textbf{82.1}\text{\scriptsize$\pm$10.5}&\textbf{79.1}\text{\scriptsize$\pm$0.3}&\textbf{63.5}\text{\scriptsize$\pm$0.0}&\textbf{64.6}\text{\scriptsize$\pm$2.1}&\textbf{73.2}\text{\scriptsize$\pm$1.5}&\textbf{83.0}\text{\scriptsize$\pm$0.0}&\textbf{78.2}\text{\scriptsize$\pm$0.7}&\textbf{85.6}\text{\scriptsize$\pm$0.9}&\textbf{32.8}\text{\scriptsize$\pm$0.6}\\

 &$\boldeq{b}_q$ &80.8\text{\scriptsize$\pm$0.4}&63.4\text{\scriptsize$\pm$1.0}&67.8\text{\scriptsize$\pm$1.5}&65.2\text{\scriptsize$\pm$0.4}&61.5\text{\scriptsize$\pm$0.8}&59.3\text{\scriptsize$\pm$0.2}&58.1\text{\scriptsize$\pm$1.8}&\textbf{83.0}\text{\scriptsize$\pm$0.0}&77.0\text{\scriptsize$\pm$0.5}&74.6\text{\scriptsize$\pm$0.6}&27.0\text{\scriptsize$\pm$0.6} \\
 &$\boldeq{b}_k$&61.2\text{\scriptsize$\pm$0.0}&54.8\text{\scriptsize$\pm$0.0}&50.0\text{\scriptsize$\pm$0.0}&61.2\text{\scriptsize$\pm$1.2}&37.5\text{\scriptsize$\pm$0.0}&51.2\text{\scriptsize$\pm$0.0}&44.5\text{\scriptsize$\pm$0.7}&82.0\text{\scriptsize$\pm$0.0}&76.4\text{\scriptsize$\pm$0.6}&38.0\text{\scriptsize$\pm$0.9}&15.6\text{\scriptsize$\pm$0.8} \\
\bottomrule[2pt]
\end{tabular}}
\caption{Downstream performance (\%) (mean$\pm$std over three runs with different random seeds).}
\label{tab:appendix_multi-seed_table_5}
\end{table*}

\newpage
\subsection{Stability Across Random Seeds}
\label{sec:appendix_multi-seed}

\begin{table}[!ht]
\centering
\begin{tabular}{cccc}
\toprule[2pt]
$\boldeq{b}_\mathcal{T}$ &  SST-2 & CoLA & STS-B \\
\midrule[1pt]
\rowcolor{mygreen!40} $\boldeq{b}_v$ & \textbf{83.9}\text{\scriptsize$\pm$1.39} & \textbf{40.2}\text{\scriptsize$\pm$2.85}& \textbf{75.3}\text{\scriptsize$\pm$0.65}\\ 
$\boldeq{b}_q$ &  75.6\text{\scriptsize$\pm$4.36}& 27.5\text{\scriptsize$\pm$3.98}&67.5\text{\scriptsize$\pm$0.28}\\ 
$\boldeq{b}_k$ &  72.6\text{\scriptsize$\pm$4.61}&4.1\text{\scriptsize$\pm$2.93}&65.6\text{\scriptsize$\pm$0.48}  \\ 
\bottomrule[2pt]
\end{tabular}
\caption{Downstream performance (\%) (mean$\pm$std over three runs with different random seeds) of fine-tuning different bias terms on representative tasks as presented in Fig. \ref{fig:rank_comparison}, with BERT$_{\mathrm{BASE}}$ in low-data regime.}
\label{tab:appendix_multi-seed_table_1}
\end{table}

\begin{table}[!ht]
\centering
\begin{tabular}{cccc}
\toprule[2pt]
Models & $\boldeq{b}_\mathcal{T}$   & SST-2 & WiC  \\
\midrule[1pt]
\rowcolor{mygreen!40} \cellcolor{white} \multirow{3}{*}{RoBERTa$_{\mathrm{BASE}}$}&
 $\boldeq{b}_v$ & \textbf{89.6}\text{\scriptsize$\pm$0.72} & \textbf{61.8}\text{\scriptsize$\pm$1.12}\\ 
&$\boldeq{b}_q$ &  84.7\text{\scriptsize$\pm$3.30}& 56.5\text{\scriptsize$\pm$0.94}\\ 
&$\boldeq{b}_k$ &  72.6\text{\scriptsize$\pm$7.84}&56.6\text{\scriptsize$\pm$0.95}  \\
\hline
\rowcolor{mygreen!40} \cellcolor{white} \multirow{3}{*}{BERT$_{\mathrm{LARGE}}$}&
 $\boldeq{b}_v$ & \textbf{89.8}\text{\scriptsize$\pm$0.58} & \textbf{67.3}\text{\scriptsize$\pm$1.03}\\ 
&$\boldeq{b}_q$ & 83.6\text{\scriptsize$\pm$0.33}& 62.6\text{\scriptsize$\pm$1.06}\\ 
&$\boldeq{b}_k$ &  73.2\text{\scriptsize$\pm$3.60}&61.5\text{\scriptsize$\pm$0.51}  \\ 
\bottomrule[2pt]
\end{tabular}
\caption{Downstream performance (\%) (mean$\pm$std over three runs with different random seeds) on SST-2 (GLUE) and WiC (SuperGLUE), with RoBERTa$_{\mathrm{BASE}}$ and BERT$_{\mathrm{LARGE}}$ in low-data regime.}
\label{tab:appendix_multi-seed_table_3}
\end{table}

\subsection{\gls{LoRA}/\gls{VeRA}/\gls{DoRA} for Bias Terms}
\label{sec:appendix_lora_vera_dora}

For a pre-trained bias vector $\boldeq{b}_{\mathcal{T},0} \in \mathbb{R}^{1\times d}$, the fine-tuned $\boldeq{b}_{\mathcal{T}}^{\prime}$ with \gls{LoRA} is presented:
\begin{align}
\boldeq{b}_{\mathcal{T}}^{\prime} = \boldeq{b}_{\mathcal{T},0}+ \Delta\boldeq{b}_{\mathcal{T}}=\boldeq{b}_{\mathcal{T},0}+\mathrm{vec}(\underline{\boldeq{B}\boldeq{A}}),
\nonumber
\end{align}
where $\boldeq{B} \in  \mathbb{\mathbb{R}}^{q\times r}$ and $\boldeq{A} \in \mathbb{\mathbb{R}}^{r\times k}$ and $r<  \mathrm{min}(q,k)$, as well as $q\times k = d$; $\boldeq{B}$ and $\boldeq{A}$ are the trainable parameters. Moreover, $\mathrm{vec}(\boldeq{B}\boldeq{A})$ means its vectorization, i.e., the operation $(\boldeq{\mathrm{B}}@\boldeq{\mathrm{A}}).\mathrm{flatten()}$ in PyTorch.
The fine-tuned $\boldeq{b}_{\mathcal{T}}^{\prime}$ with \gls{VeRA} is presented as follow:
\begin{align}
\boldeq{b}_{\mathcal{T}}^{\prime} = \boldeq{b}_{\mathcal{T},0}+ \Delta\boldeq{b}_{\mathcal{T}}=\boldeq{b}_{\mathcal{T},0}+\mathrm{vec}(\underline{\Lambda_{\boldeq{i}}} \boldeq{B}\underline{\Lambda_{\boldeq{j}}}\boldeq{A}),
\nonumber
\end{align}
where $\boldeq{i} \in \mathbb{R}^{1\times q}$ and $\boldeq{j} \in \mathbb{R}^{1\times r}$ are trainable, and formally denoted by diagonal matrices $\Lambda_{\boldeq{i}}$ and $\Lambda_{\boldeq{j}}$. Moreover, to explore the extreme parameter efficiency, we exploit the variant of \gls{VeRA}, namely \gls{VeRA}$_{1d}$, directly applied to the $\boldeq{b}_{\mathcal{T},0}$, which is $\boldeq{b}_{\mathcal{T}}^{\prime} =\boldeq{b}_{\mathcal{T},0}+\underline{\lambda}\cdot \boldeq{b}_{\mathcal{T},0}.$
$\lambda$ is a trainable scalar parameter.

\begin{table}[!ht]
\centering
\begin{tabular}{cc}
\toprule[2pt]
$\boldeq{b}_\mathcal{T}$   &+\gls{VeRA}$_{1d}$   \\ 
\midrule[1pt]
\cellcolor{mygreen!40}$\boldeq{b}_v$  &\cellcolor{mygreen!40}\textbf{80.0} \\
$\boldeq{b}_q$  &  79.5  \\
$\boldeq{b}_k$  &79.5 \\
\hline
$\Delta$Params$\downarrow$& 1.3\textperthousand  \\
\bottomrule[2pt]
\end{tabular}
\caption{\aclbaichuan{The downstream performance (\%) of variants of \gls{VeRA}$_{1d}$ designed for \gls{BEFT} on SST-2 with BERT$_{\mathrm{BASE}}$ , where $\boldeq{b}_v$ surpasses $\boldeq{b}_q$ and $\boldeq{b}_k$.}}
\label{tab:vera_1d}
\end{table}

The fine-tuned $\boldeq{b}_{\mathcal{T}}^{\prime}$ with \gls{DoRA} is presented as follow:
\begin{align}
\boldeq{b}_{\mathcal{T}}^{\prime} = \underline{m} \frac{\boldeq{v}+\Delta\boldeq{v}}{\left \| \boldeq{v}+\Delta\boldeq{v} \right \|_2} =\underline{m} \frac{\boldeq{b}_{\mathcal{T},0}+\mathrm{vec}(\underline{\boldeq{B}\boldeq{A})}}{\left \| \boldeq{b}_{\mathcal{T},0}+\mathrm{vec}(\underline{\boldeq{B}\boldeq{A}}) \right \|_2},
\nonumber
\end{align}
where $m = \left \| \boldeq{b}_{\mathcal{T}} \right \|_2$ is the magnitude scalar parameter. $\boldeq{v} \in \mathbb{R}^{1\times d}$ is the directional vector and $\boldeq{v}/\left \| \boldeq{v} \right \|_2$ is a unit vector. $\Delta\boldeq{v}$ is the directional update learned by multiplying $\boldeq{B}$ and $\boldeq{A}$. $m$, $\boldeq{B}$ and $\boldeq{A}$ are the trainable parameters.

\begin{table}[!ht]
\centering
\resizebox{\columnwidth}{!}{
\begin{tabular}{cccccc}
\toprule[2pt]
 \multirow{2}{*}{Models} & \multirow{2}{*}{$(q,k,r)$}  & Params & \cellcolor{mygreen!40} $\boldeq{b}_v$ & $\boldeq{b}_q$ & $\boldeq{b}_k$\\
 \cline{4-6}
 &&Saving&\multicolumn{3}{c}{+\gls{LoRA}}\\
\midrule[1pt]
BERT$_{\mathrm{BASE}}$ & (24,32,8) & 42\% & \cellcolor{mygreen!40}\textbf{85.0}  & 82.6 & 76.9 \\
RoBERTa$_{\mathrm{BASE}}$ &  (32,24,8) & 42\% & \cellcolor{mygreen!40}\textbf{88.0} & 86.0 & 79.5 \\
BERT$_{\mathrm{LARGE}}$ & (32,32,8) & 50\%  & \cellcolor{mygreen!40}\textbf{88.9} & 77.1  & 69.3\\
\bottomrule[2pt]
\end{tabular}}
\caption{\aclbaichuan{The downstream performance (\%) of \gls{BEFT} with \gls{LoRA} on SST-2, where $\boldeq{b}_v$ surpasses $\boldeq{b}_q$ and $\boldeq{b}_k$.}}
\label{tab:lora_beft}
\end{table}

\subsection{PiSSA and BA-LoRA for Bias Terms}
\label{sec:appendix_pissa}
We also performed extra experiments for our \gls{BEFT}+PiSSA \cite{meng2024pissa}, and our \gls{BEFT}+BA-LoRA \cite{chang2026balora} with RoBERTa$_{\mathrm{BASE}}$ on SST-2 low data regime, which highlights our key finding, i.e., $\boldeq{b}_v$ allows more effective fine-tuning than $\boldeq{b}_q$ and $\boldeq{b}_k$. In PiSSA and BA-LoRA, we reshape the bias terms from $\mathbb{R}^{1\times d}$ to $ \mathbb{R}^{q\times k}$ (where $q\times k=d$) and then apply SVD to the bias terms and only fine-tune the first $r$ rank components in two matrices ($\boldeq{A}$ and $\boldeq{B}$ format). Our results indicate that in \gls{BEFT}+PiSSA \cite{meng2024pissa}, $\boldeq{b}_v$ achieves 15.7\% higher accuracy compared to $\boldeq{b}_q$ and 22.4\% higher accuracy than $\boldeq{b}_k$. In addition, in \gls{BEFT}+BA-LoRA \cite{chang2026balora}, $\boldeq{b}_v$ achieves 13.9\% higher accuracy compared to $\boldeq{b}_q$ and 21.6\% higher accuracy than $\boldeq{b}_k$.

\color{black}

\subsection{Comparison with More Advanced \gls{PEFT} Methods}
\label{sec:appendix_sota_peft}

\begin{table}[!ht]
\centering
\resizebox{\columnwidth}{!}{
\begin{tabular}{ccccc}
\toprule[2pt]
 \multirow{2}{*}{OPT-1.3B}  & \multirow{2}{*}{Params} &SST-2  &ReCoRD& SQuAD\\
 &&classification&multiple choice&generation\\
\midrule[1pt]
\rowcolor{mygreen!40}
$\boldeq{b}_v$ & 0.04\textperthousand & 93.1  &71.5& 79.8\\
\gls{LoRA} & 1.2\textperthousand & 93.6&71.0& 79.7\\
\gls{VeRA} & 0.07\textperthousand & 92.2& 71.7& 77.6 \\ 
\gls{DoRA} & 1.3\textperthousand & 93.8  &71.8& 80.9\\ 
\bottomrule[2pt] 
\end{tabular}}
\caption{\aclbaichuan{Downstream performance (\%) of our \gls{BEFT} compared to more advanced techniques such as \gls{VeRA} \cite{kopiczko2024vera} and \gls{DoRA} \cite{liu2024dora}. Fine-tuning $\boldeq{b}_v$ requires 1.75$\times$ and 32.5$\times$ fewer parameters than \gls{VeRA} and \gls{DoRA}, respectively, while maintaining competitive downstream performance.}}
\label{tab:sota_peft}
\end{table}

\subsection{Extended to Multilingual and Commonsense Reasoning Datasets}
\label{sec:appendix_more_data}
For the multilingual dataset, we extract three different language sets (‘en’: English, ‘zh’: Chinese, ‘fr’: French) from PAWS-X data, with the XLM-RoBERTa model. In low-data regime, we consider 3333 training samples for each language set and a total of 1998 test samples from different language sets. Our results indicate that for PAWS-X datasets, fine-tuning $\boldeq{b}_v$ achieves over 9.1\% higher accuracy, compared to $\boldeq{b}_q$ or $\boldeq{b}_k$. In addition, we have performed initial experiments on CommonsenseQA, which is a question answering dataset for commonsense reasoning, and observed that fine-tuning $\boldeq{b}_v$ achieves over 17.6\% higher accuracy, compared to $\boldeq{b}_q$ or $\boldeq{b}_k$.

\color{black}
\subsection{Adding Bias Terms to More Bias-Free \glspl{LLM}}
\label{sec:appendix_more_llm_beft}

GPT-J-6B \cite{gpt-j} does not include bias terms in its attention module, and DeepSeek-Coder-Base-1.3B \cite{guo2024deepseek} does not include bias terms in the whole model.

\begin{table}[!ht]
\centering
\resizebox{\columnwidth}{!}{
\begin{tabular}{cccc}
\toprule[2pt]
Bias-Free \glspl{LLM} & \cellcolor{mygreen!40} Adding $\boldeq{b}_v$ & Adding $\boldeq{b}_q$ & Adding $\boldeq{b}_k$\\
\midrule[1pt]
DeepSeek-Coder-Base-1.3B & \cellcolor{mygreen!40} \textbf{76.9} & 67.4 & 60.3 \\ %lr=1e-2
GPT-J-6B & \cellcolor{mygreen!40} \textbf{92.8} & 88.3 & 63.8 \\ %lr=1e-3
\bottomrule[2pt]
\end{tabular}}
\caption{\aclbaichuan{Downstream performance (\%) on SST-2 dataset by adding bias terms into bias-free \glspl{LLM}, showing that for bias-free \glspl{LLM} in the low-data regime, \textit{directly adding $\boldeq{b}_v$} and fine-tuning $\boldeq{b}_v$ is sufficient without requiring any post-hoc evaluation}.}
\label{tab:bias_free_beft}
\end{table}

\section{Experimental Details}
\label{sec:appendix_hyperparameters}

We first define different training data regimes on the GLUE benchmark in Table \ref{tab:appendix_data_split}. Then we present the hyperparameters for BERT$_\mathrm{BASE}$ on the small dataset from the GLUE benchmark in Table \ref{tab:appendix_bertbase_hyperparameters} and demonstrate hyperparameters for BERT$_\mathrm{BASE}$ on the large dataset from the GLUE benchmark in Table \ref{tab:appendix_bertbase_hyperparameters_bigdata}. Next, we show the hyperparameters for BERT$_{\mathrm{BASE}}$ on WiC and CB datasets from the SuperGLUE benchmark on the low-data regime in Table \ref{tab:appendix_bertbase_hyperparameters_superglue}. Meanwhile, we show the hyperparameters for RoBERTa$_{\mathrm{BASE}}$ and BERT$_{\mathrm{LARGE}}$ on SST-2 and RTE datasets from GLUE benchmark, and WiC and CB from SuperGLUE benchmark on the low-data regime in Table \ref{tab:appendix_bertlarge_hyperparameters}. Finally, we present the hyperparameters for OPT-1.3B and OPT-6.7B in Table \ref{tab:appendix_opt1.3B_hyperparameters}.

\begin{table}[H]
\centering
\resizebox{\columnwidth}{!}{
\begin{tabular}{cccc}
\toprule[2pt]
&Low-Data &Medium-Data  &High-Data \\
\midrule[1pt]
SST-2 & 1000 & 5000 & 11000 \\
RTE & 300 & 900 & 1200  \\
QQP  & 1000 & 5000 & 9000  \\
QNLI  & 1000 & 5000 &  9000 \\
MNLI$_m$ & 1000 & 5000 &  9000\\
MNLI$_{mm}$  & 1000 & 5000 &  9000\\
CoLA  & 1710 & 3420  & 4275 \\
MRPC  & 367 & 1191& 1835\\
STS-B & 287 & 1150 & 2300 \\
\bottomrule[2pt]
\end{tabular}}
\caption{The definition of different training data regimes on the GLUE benchmark.}
\label{tab:appendix_data_split}
\end{table}

\begin{table}[H]
\centering
\resizebox{\columnwidth}{!}{
\begin{tabular}{ccccccccc}
\toprule[2pt]

& Train &Val  &\multirow{2}{*}{lr}  &Batch &\multicolumn{3}{c}{Train Epoch} &\multirow{2}{*}{Metrics} \\
&Size&Size&&Size&$\boldeq{\theta}$ &$\boldeq{b}_{\mathrm{all}}$  & $\boldeq{b}_{\mathcal{T}}$  & \\
\midrule[1pt]

SST-2 & 67k&872& 4e-4&16& 16&16&16& Accuracy \\
RTE  & 2.5k&277&1e-3& 16& 30&30&30& Accuracy\\
CoLA  &8.5k&1k&7e-4&16&16&16&16& MCC\\
MRPC  &3.7k&408&7e-4&16&16&16&60& F1\\
STS-B &5.7k&1.5k&1e-4&16&16&30&60& SCC\\
\bottomrule[2pt]
\end{tabular}}
\caption{Hyperparameters for BERT$_\mathrm{BASE}$ on small dataset from GLUE benchmark (MCC denotes Matthews Correlation Coefficient; SCC denotes Spearman Correlation Coefficient).}
\label{tab:appendix_bertbase_hyperparameters}
\end{table}

\begin{table*}[!ht]
\centering
\begin{tabular}{cccccccc}
\toprule[2pt]

& Train &Val  &\multirow{2}{*}{lr}  &Batch &\multicolumn{2}{c}{Train Epoch} &\multirow{2}{*}{Metrics} \\
&Size&Size&& Size&$\boldeq{b}_{\mathrm{all}}$  & $\boldeq{b}_{\mathcal{T}}$  & \\
\midrule[1pt]

QQP  & 364k&40k&4e-4 &16&16 &30& F1\\
QNLI  & 105k&5.5k& 1e-4&16&16& 30& Accuracy\\
MNLI$_m$  &393k&9.8k&1e-4&16&16&30& MA\\
MNLI$_{mm}$  &393k&9.8k&1e-4&16&16&30&MMA \\
\bottomrule[2pt]
\end{tabular}
\caption{Hyperparameters for BERT$_\mathrm{BASE}$ on large dataset from GLUE benchmark (MA denotes Matched Accuracy; MMA denotes Mismatched Accuracy).}
\label{tab:appendix_bertbase_hyperparameters_bigdata}
\end{table*}

\begin{table*}[!ht]
\centering
\begin{tabular}{cccccccc}
\toprule[2pt]

& Low-Data &Val  &\multirow{2}{*}{lr}  &Batch &\multicolumn{2}{c}{Train Epoch} &\multirow{2}{*}{Metrics} \\
&Regime&Size&&Size &$\boldeq{b}_{all}$& $\boldeq{b}_{\mathcal{T}}$  & \\
\midrule[1pt]
WiC  & 1000 &638& 1e-3&16&16& 16& Accuracy\\
CB &250&56&1e-3&16 &16& 60 & F1 (macro) \\
\bottomrule[2pt]
\end{tabular}
\caption{Hyperparameters for BERT$_{\mathrm{BASE}}$ on WiC and CB datasets from SuperGLUE benchmark on low-data regime.}
\label{tab:appendix_bertbase_hyperparameters_superglue}
\end{table*}

\begin{table*}[!ht]
\centering
\begin{tabular}{ccccccc}
\toprule[2pt]

& Low-Data &Val  &\multirow{2}{*}{lr}  &Batch &Train Epoch &\multirow{2}{*}{Metrics} \\
&Regime&Size&& Size& $\boldeq{b}_{\mathcal{T}}$  & \\
\midrule[1pt]

SST-2 & 1k&872& 4e-4 &16 &30& Accuracy\\
RTE  & 300 &277& 1e-3&16& 30& Accuracy\\
WiC  & 1000 &638& 1e-3&16& 30& Accuracy\\
CB &250&56&1e-3&16 & 60 & F1 (macro) \\
\bottomrule[2pt]
\end{tabular}
\caption{Hyperparameters for RoBERTa$_{\mathrm{BASE}}$ and BERT$_{\mathrm{LARGE}}$ on SST-2 and RTE datasets from GLUE benchmark, and WiC and CB from SuperGLUE benchmark on low-data regime.}
\label{tab:appendix_bertlarge_hyperparameters}
\end{table*}

\begin{table*}[!ht]
\centering
\begin{tabular}{cccccc}
\toprule[2pt]

&Low-Data Regime &Val Size& Model Adaptation & Learning Rate (lr)  &Metrics \\
\midrule[1pt]

SST-2 & 1k&872 & $\boldeq{b}_v$/$\boldeq{b}_q$/$\boldeq{b}_k$/LoRA$^{*}$/Prefix & 1e-3& Accuracy \\
\hline
RTE  &1k &277&$\boldeq{b}_v$/$\boldeq{b}_q$/$\boldeq{b}_k$/LoRA$^{*}$/Prefix & 1e-3 & Accuracy\\
\hline
BoolQ  &1k &1k&$\boldeq{b}_v$/$\boldeq{b}_q$/$\boldeq{b}_k$/LoRA$^{*}$/Prefix & 1e-3 & Accuracy\\
\hline
MultiRC  &1k &1k&$\boldeq{b}_v$/$\boldeq{b}_q$/$\boldeq{b}_k$/LoRA$^{*}$/Prefix & 1e-3 & Accuracy\\
\hline
WSC &554 &104&$\boldeq{b}_v$/$\boldeq{b}_q$/$\boldeq{b}_k$/LoRA/Prefix & 1e-2 &Accuracy\\
\hline
CB & 250&56 & $\boldeq{b}_v$/$\boldeq{b}_q$/$\boldeq{b}_k$/LoRA$^{*}$/Prefix & 1e-2& Accuracy \\
\hline
WiC & 1k&638 & $\boldeq{b}_v$/$\boldeq{b}_q$/$\boldeq{b}_k$/LoRA$^{*}$/Prefix & 1e-3& Accuracy \\
\hline
COPA & 400&100 & $\boldeq{b}_v$/$\boldeq{b}_q$/$\boldeq{b}_k$/LoRA/Prefix & 1e-4& Accuracy \\
\hline
ReCoRD & 1k&1k & $\boldeq{b}_v$/$\boldeq{b}_q$/$\boldeq{b}_k$/LoRA/Prefix & 1e-4& Accuracy \\
\hline
SQuAD &1k&1k & $\boldeq{b}_v$/$\boldeq{b}_q$/$\boldeq{b}_k$/LoRA/Prefix & 1e-3& F1 \\
\hline
DROP & 1k&1k & $\boldeq{b}_v$/$\boldeq{b}_q$/$\boldeq{b}_k$/LoRA/Prefix & 1e-3& F1\\
\bottomrule[2pt]
\end{tabular}
\caption{Hyperparameters for OPT-1.3B and OPT-6.7B (default epoch is 5; default batch size is 8; default max length the model can take is 2048, while 1536 for OPT-6.7B on DROP dataset, and small batch size with gradient accumulation is adopted for OPT-6.7B \cite{marek2025small} due to the Out-of-Memory (OOM) problem). \aclbaichuan{For LoRA$^{*}$, the learning rate is 1e-4.}}
\label{tab:appendix_opt1.3B_hyperparameters}
\end{table*}

\end{document}